\documentclass{article}
\usepackage[preprint]{corl_2025}
\usepackage{xcolor}
\usepackage{amssymb}
\usepackage{graphicx}
\usepackage{svg}
\usepackage{arydshln}
\usepackage{booktabs}
\usepackage{multirow}
\usepackage{makecell}
\usepackage{algorithm}
\usepackage{algpseudocode}
\usepackage{amsmath}
\usepackage{subfig}
\usepackage{tikz}
\usepackage{array}
\usepackage{caption}
\renewcommand{\arraystretch}{0.8}
\usepackage{enumitem}
\usepackage{titlesec}
\usepackage[hang,flushmargin]{footmisc}
\usepackage{wrapfig}

\titlespacing*{\section}{0pt}{1.2ex plus 0.5ex minus .2ex}{0.8ex plus .2ex}
\titlespacing*{\subsection}{0pt}{1ex plus 0.5ex minus .2ex}{0.5ex plus .2ex}

\title{Do LLM Modules Generalize? A Study on Motion Generation for Autonomous Driving}

\author{%
  Mingyi Wang\textsuperscript{1,2,3}\footnotemark[1]%
  \quad
  Jingke Wang\textsuperscript{2}\footnotemark[1]
  \quad
  Tengju Ye\textsuperscript{2}
  \quad
  Junbo Chen\textsuperscript{2}\footnotemark[2]%
  \quad
  Kaicheng Yu\textsuperscript{1}\footnotemark[2]
  \\
  \textsuperscript{1}Autolab, Westlake University
  \quad
  \textsuperscript{2}UDEER.AI
  \quad
  \textsuperscript{3}Zhejiang University
  \\[0.2em] 
  \texttt{wangmingyi@zju.edu.cn}, \\
  \texttt{\{jingke, tengju, junbo\}@udeer.ai}, \\[0.1em]
  \texttt{kyu@westlake.edu.cn}
}

\begin{document}
\renewcommand{\thefootnote}{\fnsymbol{footnote}}
\maketitle

\vspace{-2em}

\begingroup
\renewcommand\thefootnote{}
\footnotetext{%
\textsuperscript{*}~These authors contributed equally.\quad
\textsuperscript{\textdagger}~Co-corresponding authors. \quad
\textsuperscript{}~Accepted at CoRL 2025.%
}
\footnotetext{The project page and code can be found at: \href{https://wizard-wang01.github.io/LLM2AD/}{https://wizard-wang01.github.io/LLM2AD/}}
\endgroup

\begin{abstract}
Recent breakthroughs in large language models (LLMs) have not only advanced natural language processing but also inspired their application in domains with structurally similar problems—most notably, autonomous driving motion generation. Both domains involve autoregressive sequence modeling, token-based representations, and context-aware decision making, making the transfer of LLM components a natural and increasingly common practice. However, despite promising early attempts, a systematic understanding of which LLM modules are truly transferable remains lacking. In this paper, we present a comprehensive evaluation of five key LLM modules—tokenizer design, positional embedding, pre-training paradigms, post-training strategies, and test-time computation—within the context of motion generation for autonomous driving. Through extensive experiments on the Waymo Sim Agents benchmark, we demonstrate that, when appropriately adapted, these modules can significantly improve performance for autonomous driving motion generation. In addition, we identify which techniques can be effectively transferred, analyze the potential reasons for the failure of others, and discuss the specific adaptations needed for autonomous driving scenarios. We evaluate our method on the Sim Agents task and achieve competitive results. 
\end{abstract}



\section{Introduction}
In parallel with the recent breakthroughs in large language models~\cite{shao2024deepseekmath,touvron2023llama}, Transformer-based approaches have also achieved widespread success in the domain of autonomous driving motion generation, including tasks such as trajectory prediction~\cite{gao2020vectornet,zhou2023qcnext,shi2022motion}, traffic simulation~\cite{zhang2023trafficbots,smart,kigras}, and ego-vehicle planning~\cite{cheng2024pluto,dauner2023parting,zhang2025carplanner}. However, most prior works have focused on empirically transferring individual modules from LLMs to autonomous driving applications.
In contrast, this paper takes a more holistic perspective to explore the striking similarities between LLMs and autonomous driving motion generation, while also highlighting the subtle differences that lie beneath these similarities.

Autonomous driving motion generation refers to the task of generating future trajectories for specified agents under a set of constraints, based on contextual information from the environment and the historical states of the agents. Transformer-based models—particularly those following the GPT architecture—have achieved considerable success in this domain in recent years~\cite{kigras,motionlm,philion2023trajeglish,behaviorgpt}. Their typical workflows encompass only a subset of the overall modeling process, consisting of the following components:
\textbf{Tokenizing}: compressing contextual and motion information into a sequence of tokens that can later be decoded into trajectories;
\textbf{Positional Embedding}: encoding spatial and temporal relationships among agents, between agents and the environment, and within each agent’s trajectory;
\textbf{Pre-training}: training the model on large-scale self-supervised datasets to learn generalizable motion generation capabilities;
\textbf{Post-training}: applying fine-tuning techniques to enforce additional constraints, such as collision avoidance or human-like behavior;
\textbf{Test-time Computing}: leveraging increased compute budgets to dynamically optimize model outputs without modifying model parameters during inference.

\clearpage
\begin{figure}[htbp]
    \vspace{-10mm}
    \centering
    \includegraphics[width=0.8\textwidth]{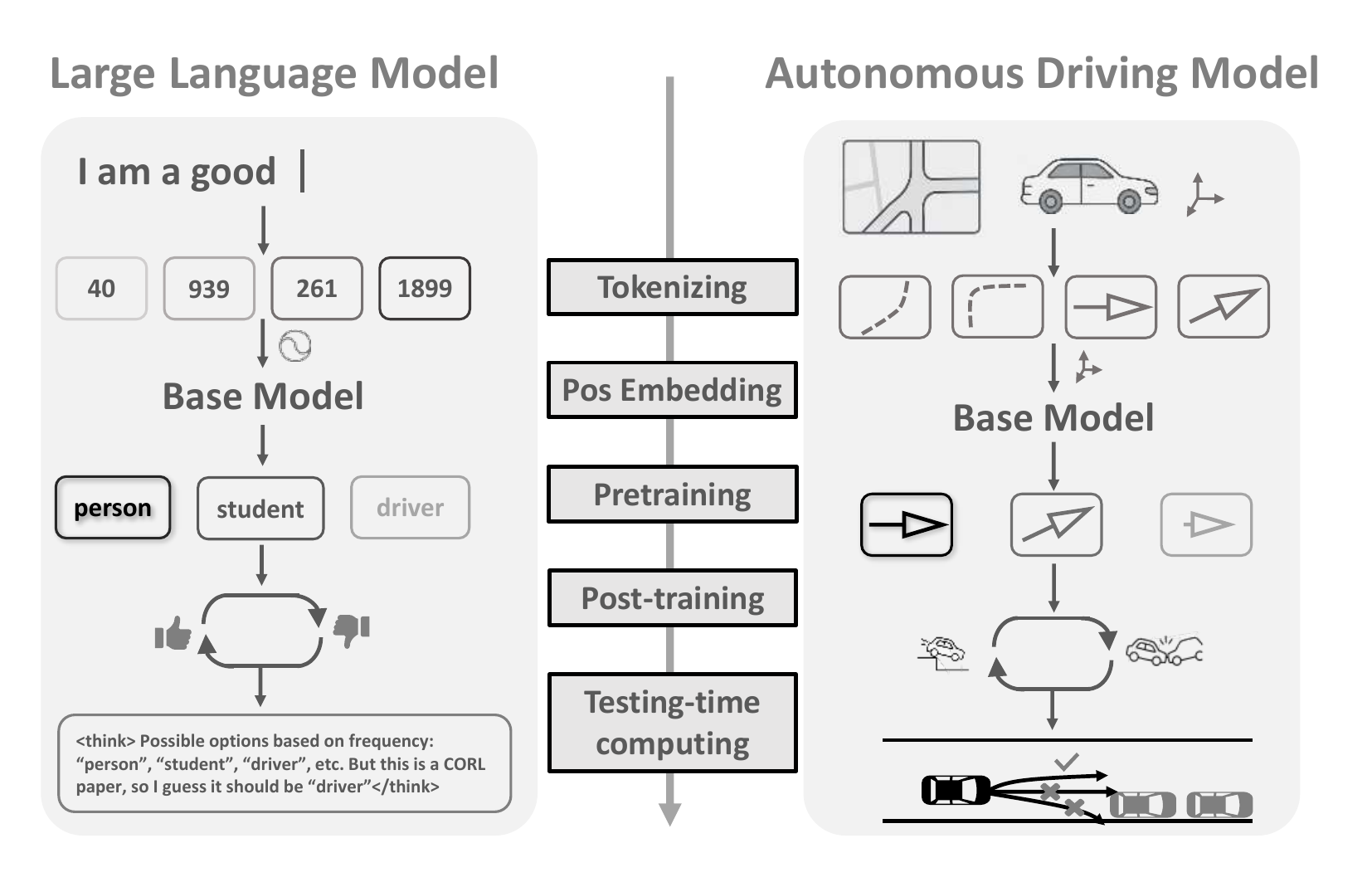}
    \caption{The autonomous driving motion generation task bears a striking resemblance to large language models.}
    \label{fig:framework}
\end{figure}

As illustrated in Figure~\ref{fig:framework}, the technical pipeline of autonomous driving motion generation exhibits a notable resemblance to that of large language models (LLMs). This observation naturally prompts the question: \textbf{which modules, proven effective in LLMs, can be directly transferred to motion generation for autonomous driving, and which ones necessitate domain-specific adaptation?} In this work, we conduct a systematic investigation of these five core components.

Our key finding is that, despite differences in application domains, several technical modules are transferable between large language models (LLMs) and motion generation tasks. As LLMs are typically pretrained with a Transformer-based autoregressive objective to capture linguistic structure, we apply the same formulation to motion generation by modeling trajectories as sequences to obtain general motion priors. Within the range of model and data scales we study, performance follows a power-law–like scaling during pretraining.
In test-time computing, LLMs often increase the sampling budget to produce multiple rollouts and then select the candidate with the highest score. The same strategy can be applied to motion generation by sampling multiple trajectories at inference and choosing the safest one under a specified risk metric.

However, we still observe important differences. According to tokenizing, language is typically composed of discrete tokens, whereas motion is inherently continuous; thus an appropriate discretization scheme is required to represent continuous trajectories with tokens. For positional encoding, beyond temporal order, spatial relationships are also crucial for motion generation, which motivates encodings that capture spatial positional structure. In post-training, techniques from LLMs can be used to induce preferences over actions; however, due to trade-offs among different reward terms, motion generation requires adapting and extending these methods.


We evaluate the transferability and necessary adaptations of various LLM-inspired modules on the Waymo Sim Agents task. Our optimized model is submitted to the Sim Agents test leaderboard, where it achieves competitive performance. The main contributions of this work are:
\begin{itemize}[left=0em, itemsep=0em, topsep=0em]
\item  To the best of our knowledge, this is the first systematic investigation into the transferability of LLM modules to to motion generation in autonomous driving.
\item We identify key factors contributing to the poor transferability of certain modules and introduce domain-specific adaptations tailored for motion generation.
\item We validate the effectiveness of the adapted modules on the Waymo Sim Agents benchmark, achieving competitive results.
\end{itemize}

\section{Autonomous Driving Motion Generation Task}
\label{sec:ad_task}
Motion generation is the task of predicting agents’ future trajectories from context and history. We focus on its application in autonomous driving. Autonomous driving motion generation is a broad concept that can be defined as generating the future trajectories $\{\hat Y_i\}_{i\in\mathcal{I}}$ of specified traffic participants $\mathcal{I}$ given the environment context information $C_t$ (map, traffic signals, etc.)  and traffic participants' historical motion $\{H_i^t\}_{i\in\mathcal{I}}$ as shown in equation~\eqref{eq:general_task}. 
\begin{equation}
\{\hat Y_i\}_{i\in\mathcal{I}} \;=\; G\bigl(C_t,\{H_i^t\}_{i\in\mathcal{I}}\bigr)
\label{eq:general_task}
\end{equation}
A wide range of tasks—including trajectory prediction, traffic simulation, and ego-vehicle planning—can be uniformly framed as specific instances of motion generation in autonomous driving. The primary differences among these tasks lie in their respective optimization objectives. For example, trajectory prediction focuses on accurately forecasting the future trajectory of a specific agent \cite{wang2022ltp,zhou2023qcnext,Min2024HierarchicalLSTM,Peng2024EfficientInteractionAware} $j$ by minimizing the deviation from the ground truth trajectory $Y_j^{\mathrm{GT}}$: $\{\hat Y_j \;=\; \arg\min_{Y}\;\bigl\lVert Y - Y_j^{\mathrm{GT}}\bigr\rVert \}$. In contrast, traffic simulation emphasizes the generation of diverse yet behaviorally plausible trajectories for all agents \cite{behaviorgpt,smart,Lucente2023BayesianMSNE}, modeled as samples from a conditional distribution: $\{Y_i\}\;\sim\;P\bigl(\{Y_i\}\mid C_t,\{H_i^t\}\bigr)$. Meanwhile, ego-vehicle planning \cite{cheng2024pluto,zhang2025carplanner,cusumano-towner2025robust,fan2024rasc} prioritizes the safety and comfort of the ego agent’s future motion, typically formulated as a constrained optimization problem: $\{\hat Y_{\mathrm{ego}} = \arg\min_{Y}\;\bigl[J_{\mathrm{safety}}(Y)+J_{\mathrm{comfort}}(Y)\bigr], \quad\mathit{s.t. vehicle\ dynamics, road\ boundaries\ constraints} \}$.

In this paper, we adopt the Waymo Sim Agents task~\cite{montali2023waymo} as the benchmark for autonomous driving motion generation, as it provides a well-balanced evaluation of the three aforementioned objectives, making it a suitable metric for assessing motion generation quality. In Sim Agents task, both training and testing scenarios are collected from real-world driving data. For any given scenario, the map and traffic light information are provided, along with a 1-second motion history of all traffic participants in the scene. The objective is to predict 8-second future trajectories for multiple specified agents (ranging from 2 to 8), including the ego vehicle. Furthermore, the model is required to generate 32 diverse future rollouts for each scenario. The benchmark includes evaluation metrics tailored to different aspects of the task: Average Displacement Error (ADE) and minimum Average Displacement Error (minADE) from rollouts for measuring prediction accuracy, realism likelihood metrics for assessing simulation quality, and collision rate, off-road rate, and motion likelihood metrics for evaluating planning performance. Further metric details can be found in the Appendix~\ref{subsubsec:sim_agents_metrics}.

\section{LLM-to-Driving Transfer Analysis}
\label{sec:llm2driving}
\subsection{Tokenizing}

Tokenization in large language models (LLMs) is typically performed at the word~\cite{bengio2003neural}, subword~\cite{kudo2018sentencepiece}, or character level~\cite{gao2020character}, each balancing vocabulary coverage and sequence length differently.
Unlike sentences in natural language, which are composed of discrete words, object motion trajectories are continuous variables. Since autoregressive models operate on discrete sequences, a discrete representation of trajectories is required. In the field of autonomous driving, an increasing number of methods~\cite{smart,motionlm,philion2023trajeglish,behaviorgpt} adopt this approach to model agent trajectories and subsequently employ next-token prediction to accomplish trajectory generation tasks.

In traditional trajectory generation tasks, encoding an agent’s historical trajectory for information fusion in Transformer-based models is often referred to as \textquotedblleft tokenization\textquotedblright~\cite{zhang2024simpl,zhou2023qcnext}, though this process is more accurately categorized as embedding, as it transforms continuous trajectory features into latent representations. In contrast, our proposed tokenization constructs a consistent, generalizable set of discrete actions to represent agent trajectories, offering better generalization across different agents and tasks.

Current trajectory tokenization paradigms can be broadly categorized into model-driven and data-driven approaches. Model-driven methods discretize control variables (e.g., acceleration, steering) and generate trajectories via kinematic models, ensuring uniform state space coverage and physical plausibility~\cite{motionlm, kigras}. Data-driven methods cluster representative trajectories into discrete tokens~\cite{smart}, and assign the nearest token during discretization. Among existing data-driven tokenization methods, the strategy adopted in SMART~\cite{smart} is the most prevalent. It employs k-disks clustering to aggregate trajectories into 1,024 representative prototypes, which are regarded as a motion vocabulary. However, it suffers from limited coverage and a strong dependence on the quality of the collected data.

Building upon this foundation, we adopt a model-driven approach to construct a motion vocabulary, following the design principle of MotionLM~\cite{motionlm}. MotionLM adopts a Verlet wrapper to tokenize trajectories by discretizing motion within the acceleration space. Based on this formulation, all agent trajectories are encoded within a global scene-centric coordinate system, a method we refer to as \textbf{Verlet-Scene}.
This encoding approach suffers from inconsistencies in the correspondence between actions and tokens, as identical agent actions can be mapped to different tokens due to variations in the coordinate system.

Therefore, to enhance token consistency across different agents, we reconstruct the motion vocabulary using an agent-centric encoding approach. The core idea is to normalize trajectory information based on the agent's current position and orientation during encoding, thereby obtaining an agent-centric representation prior to tokenization. We refer to this approach as \textbf{Verlet-Agent}. Implementation details are provided in the appendix~\ref{appendix:tokenizer_design}.

We compare two classes of tokenization methods for motion generation: the model-based \textbf{Verlet-Scene} and \textbf{Verlet-Agent}, and the data-driven method from \textbf{SMART}\cite{smart}. Both are common choices for constructing tokenizers. Keeping all other components of the baseline fixed, we ablate only the tokenizer, and we report the results in Table~\ref{tab:tokenizer}
~\footnote{Metrics in Section \ref{subsec:benchmark} are from the official Sim Agents Leaderboard. Owing to submission limits and long evaluation times, all other tables report results on the same locally sampled validation set unless noted.}.

\vspace{-5mm}
\begin{table}[htbp]
    \centering
    \caption{Tokenizer Design}
    \label{tab:tokenizer}
    \renewcommand{\arraystretch}{1.2}
    \resizebox{0.75\textwidth}{!}{
    \begin{tabular}{cccccccc}
        \toprule
        \multirow{2}{*}{\shortstack{Tokenizer\\Design}} & \multirow{2}{*}{\shortstack{Vocabulary\\Size}} & \multicolumn{3}{c}{\textbf{10\% Data}} & \multicolumn{3}{c}{\textbf{100\% Data}} \\
        \cmidrule(r){3-5} \cmidrule(l){6-8}
        & & Acc$\uparrow$ & ADE(8s)$\downarrow$ & FDE(8s)$\downarrow$ & Acc$\uparrow$ & ADE(8s)$\downarrow$ & FDE(8s)$\downarrow$ \\
        \midrule
        SMART  & 2048 & 0.349 & 4.58 & 12.76 & 0.405 & 3.53 & 9.75 \\
        Verlet-Scene & 169  & 0.414 & 4.11 & 11.26 & 0.454 & 3.20 & 8.87 \\
        Verlet-Agent & 169  & \textbf{0.437} & \textbf{3.84} & \textbf{10.66} & \textbf{0.460} & \textbf{3.13} & \textbf{8.67} \\
        \bottomrule
    \end{tabular}
    }
\end{table}
\vspace{-2mm}
Data-driven approaches typically require larger vocabularies, which naturally lead to lower token prediction accuracy compared to model-based methods. Moreover, the increased vocabulary size makes such methods more prone to out-of-distribution issues during closed-loop evaluations.
In model-based approaches, Verlet-Scene tokenizes actions from a global scene-centric perspective, introducing ambiguity, thereby complicating the model's learning process. In contrast, Verlet-Agent effectively mitigates this issue, resulting in improved performance.

\textbf{Model-based tokenizers encode the agent’s action space using a smaller vocabulary, resulting in higher classification accuracy compared to data-driven tokenizers. Additionally, compared to Verlet-Scene, Verlet-Agent provides a more consistent mapping from tokens to physical actions, leading to more precise action modeling.}


\subsection{Positional Embedding}
Transformer architectures naturally incorporate positional encoding (PE) mechanisms, primarily to capture sequential dependencies crucial for Natural Language Processing (NLP) tasks~\cite{vaswani2017attention}. In contrast, autonomous driving environments present more complex spatial topologies, where a one-dimensional PE typical in NLP is less directly applicable. In particular, the tokens representing agents are order-agnostic, and no explicit sequential relationship exists among them. Meanwhile, the topology of the map is highly complex, making it difficult to define an order for the map elements.

Early trajectory prediction models therefore often omitted positional embedding~\cite{huang2022survey}. However, relative pose relationships between agents and map elements remain critical in driving contexts. Some approaches address this by explicitly transforming coordinate systems to encode relative poses, but such methods incur significant memory overhead~\cite{zhang2024simpl,zhou2023query}.

To address this, the recently proposed Directional Relative Positional Embedding (DRoPE)~\cite{drope} technique introduces a novel way to encode relative positions by embedding agent and map information in local coordinate frames and applying Rotary Positional Embedding (RoPE)~\cite{su2024roformer} during attention. This provides a promising new perspective for positional embedding in autonomous driving models.

However, a key distinction from NLP tasks lies in the high semantic similarity among map tokens under different local coordinate frames. This similarity weakens the discrimination power of attention when applying RoPE-based methods to map features.

Motivated by this observation, we propose an enhanced Global-DRoPE method. Specifically, we maintain agent and map encodings in the global coordinate system to preserve rich semantic information, while incorporating relative positional cues during cross-instance attention following the DRoPE framework, as shown in Figure~\ref{fig:global_drope}.

Table \ref{tab:positional_embedding} presents a comparison of different positional embedding strategies under identical experimental settings. As shown, vanilla 1D PE methods, commonly used in LLMs, negatively impact performance in autonomous driving contexts, even compared to models without any PE. In contrast, applying DRoPE-based relative pose encoding achieves a lower off-road rate, attributable to enhanced fine-grained spatial reasoning. Further improvements are achieved by our Global-DRoPE variant, demonstrating the importance of preserving semantic richness alongside relative position information. Implementation details are provided in Appendix~\ref{appendix:positional_embedding}.

\textbf{In summary, we find that while DRoPE methods imported from LLMs substantially enhance spatial reasoning in autonomous driving models, their potential can be further amplified by adapting them to preserve global semantic context.}

\vspace{-2mm}
\begin{table}[htbp]
  \centering
  \begin{minipage}[t]{0.35\textwidth}
    \centering
    \includegraphics[width=0.9\linewidth]{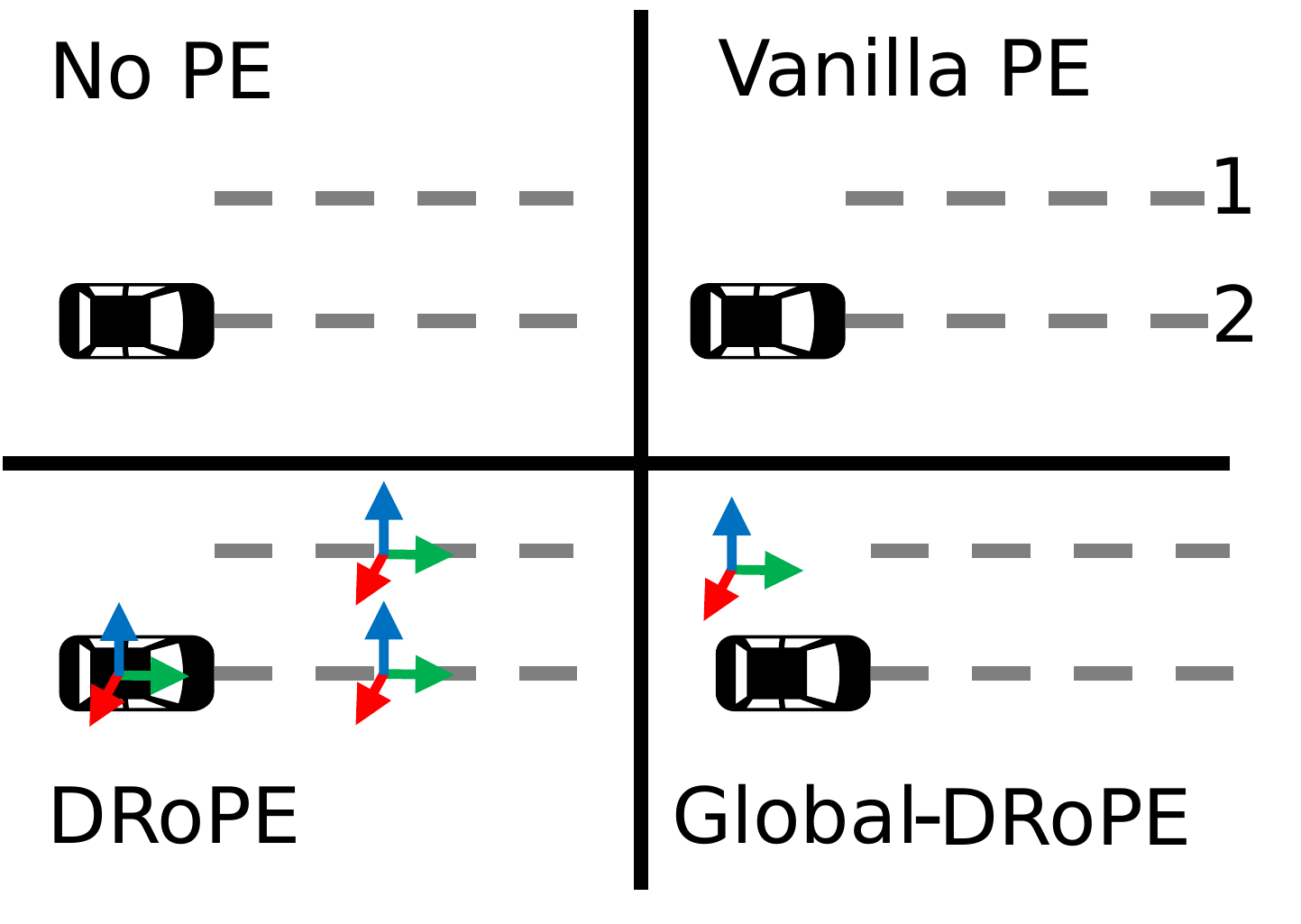}
  \end{minipage}%
  \hfill
  \begin{minipage}[t]{0.6\textwidth}
    \centering
    \vspace{-3cm}
    \renewcommand{\arraystretch}{1.2}  
    \setlength{\tabcolsep}{6pt}  
    \captionsetup{type=table}
    \captionof{table}{Comparison of different positional embedding methods on the Sim Agents task.}
    \label{tab:positional_embedding}
    \resizebox{0.95\linewidth}{!}{
      \begin{tabular}{lcccc}
        \toprule
        \textbf{Method} & \textbf{ADE}$\downarrow$ & \textbf{minADE}$\downarrow$ & \textbf{Collision Rate}$\downarrow$ & \textbf{Offroad Rate}$\downarrow$ \\
        \midrule
        No PE        & 3.13 & 1.52 & 0.146 & 0.198 \\
        Vanilla PE   & 3.15 & 1.56 & 0.140 & 0.208 \\
        DRoPE        & 3.18 & 1.49 & 0.150 & 0.186 \\
        Global-DRoPE & \textbf{3.13} & \textbf{1.45} & \textbf{0.139} & \textbf{0.175} \\
        \bottomrule
      \end{tabular}
    }
  \end{minipage}
  \captionof{figure}{Schematic diagram of different PEs. When using a local coordinate system, the lane features will be too similar.}
  \label{fig:global_drope}
\end{table}

\vspace{-5mm}
\subsection{Pre-training and Scaling Law}

Our model training is highly similar to that of LLMs, and we similarly observe that scaling laws hold to some extents. The model is trained under the next-token prediction paradigm, where it outputs the probability distribution over the next-step actions (motion tokens) of all agents. We train the model from scratch on the Waymo Open Motion Dataset, and during inference we adopt an autoregressive decoding strategy to generate motion token sequences, which are then composed into complete fine-grained trajectories.


Agent trajectories also exhibit sequential characteristics, and prior studies have shown that this paradigm can be effectively applied to trajectory modeling~\cite{philion2023trajeglish,motionlm,behaviorgpt}.
We adopt a Transformer-based decoder architecture~\cite{vaswani2017attention} for our motion generation model, similar to large language models, but with modifications to capture the attention of surrounding agents' trajectories and the associated map information. Our decoder includes four key components: (1) temporal self-attention, (2) self-attention among agents, (3) cross-attention with map features, and (4) cross-attention with other agents. The architectural design is detailed in Appendix~\ref{appd:baseline}.


In LLMs, the training data size and model parameters are crucial for generation quality~\cite{kaplan2020scaling}. Researchers have noted that the loss function follows a power-law relationship with the training data size and model parameters, a phenomenon known as the scaling law~\cite{kaplan2020scaling}. Motivated by this, we explore the applicability of the scaling law to our autoregressive motion generation model.

We train our motion generation model from scratch and test whether scaling laws apply to motion generation models. Models with varying parameter scales are trained on different sizes of training data on the Waymo train set and evaluated on the whole Waymo validation set. To enhance map-aware representations, we employ a data augmentation strategy for autoregressive models, detailed in Appendix~\ref{appd:pre-training}. This strategy increases the available training data. The full training set contains approximately 35 million tokens, with experiments using subsets of 10\%, 100\%, and 800\% (augmented) of the data. The results are summarized in Table~\ref{tab:scaling_law}.

\vspace{-3mm}
\begin{table}[htbp]
    \centering
    \renewcommand{\arraystretch}{1.2}
    \caption{Scaling Law on GPT Layers}
    \label{tab:scaling_law}
    \setlength{\tabcolsep}{2pt}
    \resizebox{0.75\textwidth}{!}{
    \begin{tabular}{cc|cccc|cccc|cccc}
        \toprule
        \multirow{2}{*}{Model} & \multirow{2}{*}{\shortstack{Paras}} & 
        \multicolumn{4}{c}{\textbf{10\% data}} & 
        \multicolumn{4}{c}{\textbf{100\% data}} & 
        \multicolumn{4}{c}{\textbf{Aug data}} \\
        \cmidrule(r){3-6} \cmidrule(r){7-10} \cmidrule(l){11-14}
        & & 
        \shortstack{Train\\Loss}$\downarrow$ & \shortstack{Val\\Loss}$\downarrow$ & \shortstack{Train\\Acc}$\uparrow$ & \shortstack{Val\\Acc}$\uparrow$ & 
        \shortstack{Train\\Loss}$\downarrow$ & \shortstack{Val\\Loss}$\downarrow$ & \shortstack{Train\\Acc}$\uparrow$ & \shortstack{Val\\Acc}$\uparrow$ & 
        \shortstack{Train\\Loss}$\downarrow$ & \shortstack{Val\\Loss}$\downarrow$ & \shortstack{Train\\Acc}$\uparrow$ & \shortstack{Val\\Acc}$\uparrow$ \\
        \midrule
        Mini & 0.8M  & 1.72 & 1.72 & 0.412 & 0.414 & 1.59 & 1.60 & 0.452 & 0.451 & 1.54 & 1.52 & 0.456 & 0.460 \\
        Medium & 3.7M  & 1.53 & 1.65 & 0.451  & 0.431 & 1.50 & 1.51 & 0.466 & 0.463 & 1.47 & \textbf{1.48} & 0.473 & 0.470 \\
        Big & 5.3M  & 1.51 & 1.66 & 0.460  & 0.431 & 1.48 & 1.50 & 0.469 & 0.464 & 1.43 & \textbf{1.48} & 0.483 & \textbf{0.473} \\
        Large & 11.6M & 1.52 & 1.64 & 0.462  & 0.432 & 1.44 & 1.49 & 0.480 & 0.469 & \textbf{1.37} & 1.49 & \textbf{0.497} & 0.472 \\
        \bottomrule
    \end{tabular}
    }
\end{table}
\vspace{-2mm}
As shown in Table~\ref{tab:scaling_law}, for models with smaller parameter sizes, increasing the volume of training data leads to lower validation loss and higher prediction accuracy, thereby improving overall model performance. Under conditions of sufficient training data (Aug data), enlarging the model size further enhances predictive accuracy, and the model exhibits behavior consistent with the scaling law within a certain range.

However, such improvements are subject to practical limitations. Due to constraints in dataset size, when the number of model parameters exceeds the size of the Big model, a 100\% increase in parameters results in only a 2\% improvement in training accuracy, accompanied by signs of overfitting even on the largest available dataset. To balance computational cost and diminishing performance gains, we select the model with approximately 5.3M parameters as the base architecture for subsequent experiments. \textbf{In summary, our autoregressive model exhibits scaling behavior: when sufficient training data is available, increasing model size yields significant performance gains, consistent with the scaling law.}


\subsection{Post-training}
Early research on large language models (LLMs) primarily focused on pre-training, with the belief that extrapolating the scaling law curve would lead to continued gains in model intelligence~\cite{wei2022emergent}. However, as model size and dataset scale increased, the rate of improvement began to diminish~\cite{kaplan2020scaling,hoffmann2022training}. This has motivated the community to explore post-training techniques as a more efficient means of enhancing model capabilities~\cite{christiano2017deep,ziegler2019fine,ouyang2022training}.

In the domain of motion generation for autonomous driving, imitation learning-based pre-training inherently suffers from causal confusion and compounding error~\cite{lu2023imitation}. Higher open-loop accuracy does not necessarily translate to safer closed-loop trajectory outcomes. Inspired by developments in LLMs, several recent works~\cite{motionlm-rl,catk,kigras,rowe2024ctrl,cusumano2025robust} in autonomous driving have proposed post-training strategies to address this issue. In this section, we provide a systematic study and evaluation of such approaches, which can be broadly categorized into imitation learning-based and reinforcement learning-based methods.

In imitation learning methods, we first focus on \textbf{Supervised Fine-tuning (SFT)}, a technique widely adopted in large language models (LLMs)~\cite{zhang2023instruction}. In the context of autonomous driving motion generation, a straightforward idea is to identify scenes where collisions occur during closed-loop evaluation as safety-critical cases, and use these cases to fine-tune the pre-trained model. 

Reinforcement learning–based post-training represents another active research direction. The most intuitive approach is to apply \textbf{REINFORCE}~\cite{williams1992simple} algorithms to fine-tune the trajectory policy based on environment feedback. However, in autonomous driving, collision or non-collision outcomes are inherently sparse and subtle, leading to high variance in the reward signal. To address this, methods such as \textbf{Advantage Actor-Critic (A2C)}~\cite{mnih2016asynchronous} are commonly employed, where a critic network estimates the value function to reduce variance through advantage estimation.
The recent success of \textbf{Group Relative Policy Optimization (GRPO)}~\cite{shao2024deepseekmath} in large model training further inspires us: group-based advantage comparisons eliminate the need for explicit value estimation, reducing memory overhead, while regularizing the KL divergence with respect to the reference policy helps stabilize RL updates.

We implement representative methods from each of the above categories and conduct a comprehensive comparison. The results are summarized in Table \ref{tab:post_training}. Experimental results show that compared to the pre-trained baseline, the improvement brought by \textbf{SFT} is very limited. This is primarily due to the scarcity of safety-critical data and the inherent limitations of imitation learning itself. In contrast, while \textbf{REINFORCE} and \textbf{A2C} methods significantly reduce collision and off-road rates, they lead to lower realism scores. This is because these methods optimize the policy conservatively, which results in substantial deviations from human-like behavior and thus lower scores. In comparison, the GRPO method achieves the best overall performance, enabling safer motion generation without significantly deviating from human strategies. Detailed implementations of these methods are provided in Appendix~\ref{appendix:post_training}.

\vspace{-5mm}
\begin{table}[htbp]
\centering
\renewcommand{\arraystretch}{1.2}
\caption{Comparison of post-training results.}
\resizebox{0.80\textwidth}{!}{
\begin{tabular}{lccccc}
\toprule
\textbf{Method} & \textbf{Realism Score (Meta)}$\uparrow$ & \textbf{ADE}$\downarrow$ & \textbf{minADE}$\downarrow$ & \textbf{Collision Rate}$\downarrow$ & \textbf{Offroad Rate}$\downarrow$ \\
\midrule
Baseline     & 0.724 & 2.82 & 1.32 & 0.124 & 0.153 \\
SFT          & 0.726 & 2.84 & 1.33 & 0.123 & 0.152 \\
REINFORCE    & 0.694 & 2.85 & 2.12 & 0.087 & 0.113 \\
A2C          & 0.703 & 2.77 & 2.01 & 0.075 & 0.107 \\
GRPO         & \textbf{0.728} & 2.59 & 1.46 & 0.093 & 0.120 \\
\bottomrule
\end{tabular}
}
\label{tab:post_training}
\end{table}
\vspace{-2mm}
\begin{figure}[ht]
    \centering
    \includegraphics[width=1.0\textwidth]{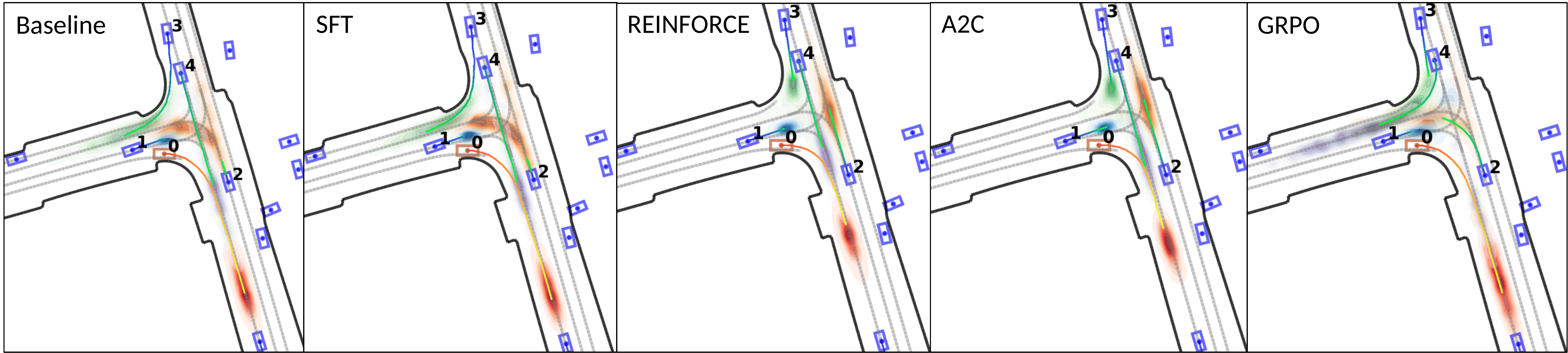}
    \caption{Comparison of different post-training methods, showing the distribution of agents’ endpoint positions at 8 seconds in 256 future rollouts.}
    \label{fig:post_training_result}
\end{figure}
\vspace{-2mm}
We conducted rollout evaluations of models trained with different post-training methods. Figure~\ref{fig:post_training_result} shows the distribution of predicted endpoints for a specific vehicle in a specific scenario. It can be observed that the baseline model produces a relatively dispersed distribution over future states, covering diverse probabilistic outcomes. However, it exhibits safety issues, with some trajectories colliding with boundaries. \textbf{The SFT policy shows only marginal improvement over the baseline, with no significant difference in behavior. In contrast, models trained with REINFORCE and A2C effectively reduce collision and off-road rates, but their predictions become overly concentrated. Reinforcement learning without constraints tends to strongly amplify the conservative modes of the pre-trained model while suppressing exploratory behaviors. The GRPO-based approach achieves a better balance between safety and human-likeness, and obtains the highest overall realism score.}




\subsection{Testing Time Computing}
Recent progress in large language models (LLMs) has sparked interest in test-time computing—a set of inference-time techniques that enhance adaptability, controllability, and efficiency without altering pre-trained weights~\cite{zarriess2021decoding}. These include test-time adaptation, constrained decoding, and search-based generation, enabling LLMs to dynamically tailor outputs to specific tasks~\cite{radford2019language,surveyofllm}. Notably, decoding-time control strategies such as top-k/top-p sampling, finite-state constrained decoding, and search-based methods enable LLMs to explore diverse outputs while applying task-specific constraints and performing re-ranking during inference~\cite{fan2018hierarchical,holtzman2019curious,zhang2019dialogpt}.

This perspective aligns naturally with motion generation for autonomous driving, where models propose candidate future motions based on multi-agent context. Although the generative backbone remains fixed, test-time post-selection—evaluating sampled trajectories with respect to cost functions like collision risk or comfort—can significantly enhance planning quality by selecting the most suitable option.

In this work, we evaluate and compare three test-time computing strategies based on trajectory clusters generated in parallel: clustering, safety filtering, and safety filtering followed by clustering. The comparative results are summarized in Table~\ref{tab:test_time_comparison}~\footnotemark[1].

\vspace{-3mm}
\begin{table}[ht]
\centering
\caption{Comparison of different test-time computing strategies}
\renewcommand{\arraystretch}{1.2}
\resizebox{0.95\textwidth}{!}{
\begin{tabular}{lccccccc}
\toprule
\makecell[c]{\textbf{Method}$\downarrow$} & \textbf{Realism.}$\uparrow$ & \textbf{Kin. Metrics}$\uparrow$ & \textbf{Inter. Metrics}$\uparrow$ & \textbf{Map Metrics}$\uparrow$ & \textbf{Collis. Rate}$\downarrow$ & \textbf{Offroad Rate}$\downarrow$ & \textbf{Runtime (s)}$\downarrow$ \\
\midrule
\makecell[c]{Baseline} & 0.724 & 0.468 & 0.745 & 0.840 & 0.124 & 0.153 & \textbf{0.69} \\
\makecell[c]{MoreRollouts + Cluster} & 0.732 & 0.466 & 0.764 & 0.844 & 0.111 & 0.159 & 5.20 \\
\makecell[c]{MoreRollouts + Search} & 0.758 & 0.466 & 0.803 & \textbf{0.864} & \textbf{0.115} & \textbf{0.052} & 7.62 \\
\makecell[c]{MoreRollouts + Search + Cluster} & \textbf{0.759} & \textbf{0.467} & \textbf{0.806} & \textbf{0.864} & 0.118 & 0.054 & 11.31 \\
\bottomrule
\end{tabular}
}

\label{tab:test_time_comparison}
\end{table}

Since our model adopts a GPT-based architecture for motion token generation, parallel rollout becomes straightforward. We duplicate the target agents $N$ times to perform parallel rollout with low computational cost, as both map information and other agents' trajectories are globally shared. This setup ensures that memory is the primary bottleneck, while computational overhead remains minimal when GPU resources are sufficient. Additionally, collision and out-of-bounds checks reuse the GPU-based environment from our reinforcement learning setup, enabling efficient post-search.

\textbf{Overall, post-search significantly reduces the collision rate of generated trajectories, and combining post-search with clustering achieves the highest weighted performance score. However, it also significantly increases computational runtime overhead. Considering both effectiveness and efficiency, using rollouts with search emerges as the more favorable choice.}


\subsection{Benchmark Result}
\label{subsec:benchmark}

\begin{wraptable}[10]{r}{0.5\textwidth} 
\centering
\vspace{-5mm} 
\renewcommand{\arraystretch}{1.2}
\setlength{\tabcolsep}{5pt}
\caption{Comparison with SOTAs in Sim Agents Leaderboard}
\resizebox{\linewidth}{!}{%
\begin{tabular}{lcccc}
\toprule
\textbf{Method} & \textbf{Realism}$\uparrow$ & \textbf{Kine.}$\uparrow$ & \textbf{Inter.}$\uparrow$ & \textbf{Map}$\uparrow$ \\
\midrule
GUMP\cite{gump}\footnotemark[2] & 0.743 & 0.478 & 0.789 & 0.740 \\
Behavior GPT\cite{behaviorgpt}\footnotemark[2] & 0.747 & 0.433 & 0.800 & 0.764 \\
SMART\cite{smart}\footnotemark[2] & 0.751 & 0.445 & 0.805 & 0.763 \\
KiGRAS\cite{kigras}\footnotemark[2] & 0.759 & 0.469 & 0.806 & 0.770 \\
DRoPE-Traj\cite{drope}\footnotemark[2] & 0.762 & 0.478 & 0.806 & 0.768 \\
UniMM\cite{unimm}\footnotemark[3] & 0.783 & 0.491 & 0.809 & 0.916 \\
Catk\cite{catk}\footnotemark[3] & \textbf{0.785} & \textbf{0.493} & \textbf{0.811} & \textbf{0.917} \\
\midrule
Combination of \\Our Ablated Modules & 0.778 & 0.487 & 0.805 & 0.911 \\
\bottomrule
\end{tabular}
}
\label{tab:meta_result}
\end{wraptable}
\vspace{-2mm}

Guided by the preceding analysis of each module, we curated an optimized configuration and submitted the resulting model to the Sim Agents Test leaderboard. We use Verlet-Agent (tokenizer) + Global-DRoPE (positional embedding) + Big Model (pre-training) + GRPO (post-training) + MoreRollouts-Search-Cluster (testing-time computing) and the detailed configuration can be found in Appendix~\ref{sec:benchmark_config}. The comparative performance against other SOTA approaches is summarized in Table~\ref{tab:meta_result}.

\footnotetext[1]{The runtime represents the mean time on a NVIDIA RTX 3060 GPU to generate 32 rollouts for one scenario.}
\footnotetext[2]{Methods marked with $^\dagger$ were evaluated in Sim Agents Challenge 2024. In 2025 their Realism Score is expected to rise by 0.014. Methods marked with $^\ddagger$ are unpublished paper method, and since the Waymo Sim Agents leaderboard is continuously updated, we report results as of April 30, 2025.}


\section{Conclusion}
\label{sec:conclusion}
In this work, we explored the transferability of various LLM modules to motion generation in autonomous driving and analyzed the reasons behind their success or failure. We validated our final integrated approach on the Sim Agents benchmark and demonstrated its effectiveness. We believe our findings offer valuable insights for the broader autonomous driving community, particularly in the areas of prediction, simulation, and planning. In future work, we aim to extend our approach to downstream planning tasks and investigate how the environment models trained in our framework can facilitate planning.

However, our study leaves several important directions unexplored. One such direction is the investigation of LLM-specific mechanisms such as Mixture-of-Experts (MoE). Moreover, due to limited data and computational resources, we have not studied the scaling behavior of larger models. Our current experiments are also restricted to vector-based environments without incorporating multi-modal sensor inputs.


\clearpage


\bibliography{example}  

\clearpage
\section*{Limitations}
\subsection{Limitations of Sim Agents Benchmark}
Overall, the Waymo Sim Agents leaderboard provides a solid testbed for comprehensive evaluation across prediction, simulation, and planning tasks. However, its primary focus remains on simulation fidelity. As a result, certain evaluation metrics—particularly the off-road likelihood—can lead to misleading assessments in specific scenarios and, to some extent, penalize RL-based methods.

\begin{figure}[htbp]
    \centering
    \includegraphics[width=0.8\textwidth]{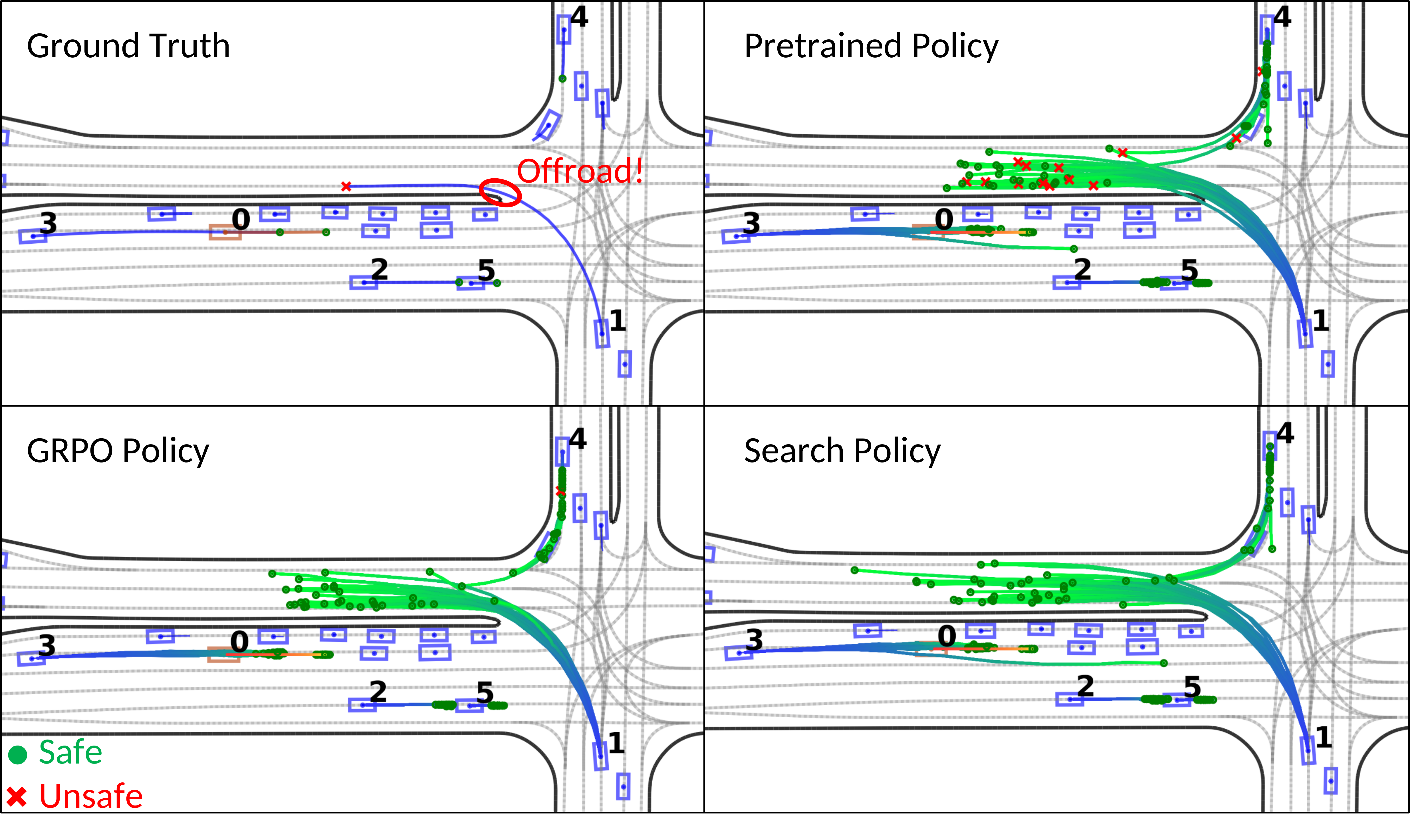}
    \caption{A limitation of using likelihood-based evaluation. Due to observation noise, the ground-truth trajectory includes a false-positive collision with the road boundary. In this case, pre-trained models that produce more collisions may achieve higher likelihood scores, as they better match the erroneous ground-truth distribution. In contrast, safer strategies, such as GRPO and search-based rollout selection, which avoid collisions, receive lower scores for deviating from the observed (but flawed) distribution.}
    \label{fig:benchmark_limitation}
\end{figure}

For instance, Figure \ref{fig:benchmark_limitation} illustrates a case in scenario \textbf{dc9b2c377e20b2be} where, due to observation noise, the ground-truth human-driven vehicle collides with the road boundary. In this situation, the scores of RL and pre-training methods are shown in the figure, revealing that the safer RL strategy receives significantly lower scores than the pre-training policy. This discrepancy arises because the evaluation emphasizes likelihood matching. The ground-truth trajectory includes a collision, while all RL-generated rollouts avoid it, thus deviating from the empirical distribution. Moreover, based on our manual review of the validation set, such cases occur in approximately $6\%$ of the scenarios, indicating that their impact on the final evaluation is non-negligible.

Although such evaluation may incentivize more “human-like” behavior in simulation tasks, it is inherently unfair to methods that achieve higher safety. Fortunately, the organizers of the Waymo Sim Agents challenge have started to recognize this issue and have indicated intentions to revise the scoring mechanism in the 2025 leaderboard. Nevertheless, since most of our evaluations were conducted under the 2024 scoring standard, our method is still affected by these limitations.

\vspace{-10mm}
\begin{table}[htbp]
\centering
\caption{Comparison of likelihood score and offroad rate in scenario \textbf{dc9b2c377e20b2be} shown in Figure~\ref{fig:benchmark_limitation}.}
\begin{tabular}{lccc}
    \toprule
    Method & Offroad Rate($\%$) $\downarrow$ & Offroad Likeli. $\uparrow$ & Realism Score$\uparrow$\\
    \midrule
    Pre-train & 6.7 & 0.771  & \textbf{0.700} \\
    GRPO     & \textbf{0.5} & \textbf{0.176}  & 0.550 \\
    Search   & \textbf{0.5} & \textbf{0.176}  & 0.548 \\
    \bottomrule
\end{tabular}
\label{tab:likelihood_vs_safety}
\end{table}

\subsection{Limitations of Our Method}
\subsubsection{Handling of Other Agents}
\vspace{-2mm}
In our method, we only include the 2 to 8 agents of interest required by Waymo for simulation in the iterative rollout process using the GPT model. Other background agents are not fed into the GPT; instead, their scene encoder outputs are passed through a cross-attention module focused on the interest agents, followed by an MLP to directly predict their trajectories over the next 8 seconds.

In fact, our method is GPU memory-efficient and could easily accommodate many more agents in parallel inference. Additionally, the use of DRoPE for positional embedding provides strong extrapolation capabilities.
However, the observation quality of “other agents” in the Waymo dataset is poor, preventing us from using them for GPT training. Trajectory data from “other agents” contains substantial noise and missing values~\cite{montali2023waymo}, which could mislead the model. Therefore, the number of agents fed into the GPT during training is limited to 2 to 8; inputting too many obstacle agents at inference time may lead to out-of-distribution issues. We believe that this limitation can be overcome by using datasets with higher data quality. In the future, we plan to explore datasets such as nuPlan and Argoverse2 to obtain more high-quality scenarios.

\subsubsection{Failed Cases}
\begin{figure}[h]
    \centering
    \includegraphics[width=0.9\textwidth]{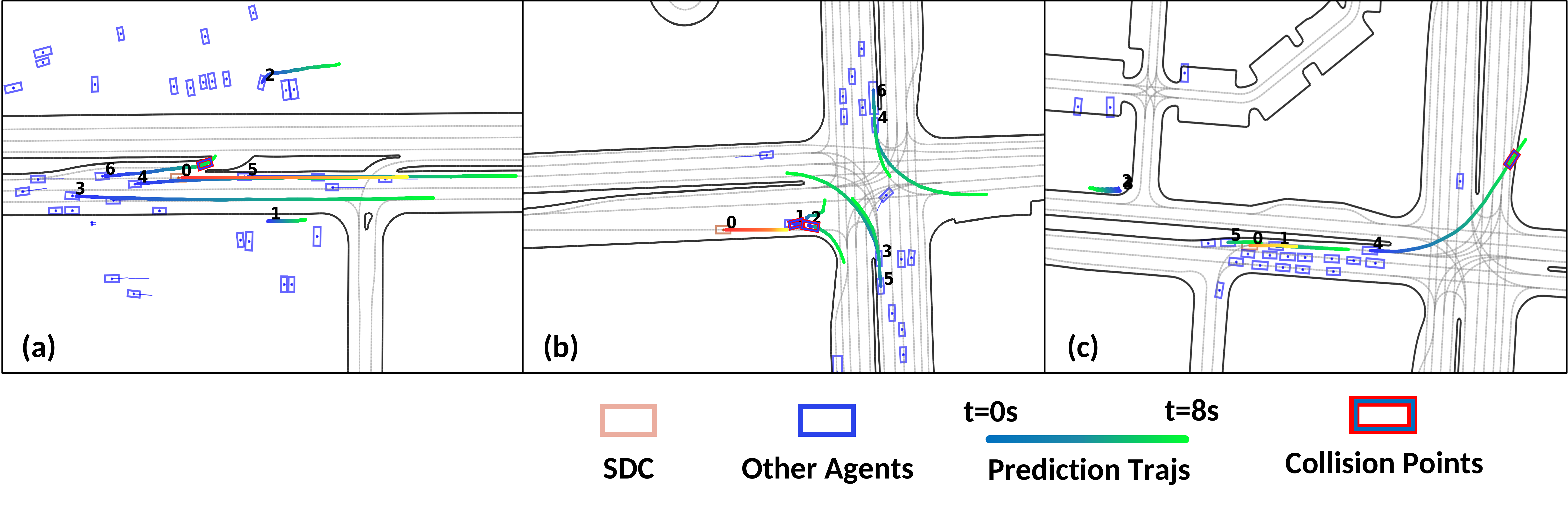}
    \caption{The limitation of our method.}
    \label{fig:method_limitation}
\end{figure}

\vspace{-2mm}
Although our method generally achieves good safety performance, there are still scenarios where collisions or unrealistic behaviors may occur. For example:

\begin{itemize}[left=0em, itemsep=0em, topsep=0em]
\item{When \textbf{the initial state is too close to a boundary} or requires a tight turn along the boundary, the model may collide with the road edge, as illustrated in Figure \ref{fig:method_limitation} (a).}
\item{When \textbf{the initial states of two agents are already in collision}, the model may encounter an out-of-distribution situation, resulting in uncontrolled rollouts, as shown in Figure \ref{fig:method_limitation} (b).}
\item{\textbf{In some simple scenarios}, the model may rarely exhibit a repetition phenomenon, similar to word repetition in large language models. In our experiments, the most frequent repeated token is $<84>$, representing repeating the previous frame’s action. Although such cases should be easy to handle, once trapped in this loop the model loses reasoning ability and may collide with obstacles or road boundaries (Figure \ref{fig:method_limitation} (c)). Even though this issue is extremely rare (under $0.5\%$ of cases), the safety risks and underlying causes merit further investigation. We hypothesize it relates to progressive out-of-distribution (OOD) drift, but the exact triggering scenarios remain open for future study.}

\end{itemize}
\vspace{-3mm}

\vspace{-3mm}
\section*{Acknowledgement}
\vspace{-3mm}
This work is partially supported by the National Natural Science Foundation of China (NSFC) under grant No. 62403389 and the Provincial Natural Science Foundation of Zhejiang under grant No. QKWL25F0301.

This work was carried out during the first author’s internship at Udeer.AI, in collaboration with Westlake University. The authors gratefully acknowledge the valuable support and collaboration provided by Udeer.AI and Westlake University.

\clearpage

\section{Appendix}
\label{sec:appendix}
\subsection{Related Work}
\label{subsec:related_work}
In recent years, large language models (LLMs) have achieved remarkable breakthroughs in natural language processing, driven by the synergistic advancement of several key technologies~\cite{surveyofllm}. These include efficient tokenization methods~\cite{xue2022byt5,tay2021charformer}, flexible positional embedding strategies~\cite{dai2019transformer,he2020deberta,su2024roformer}, large-scale autoregressive pre-training paradigms~\cite{chang2022maskgit,strudel2022self}, post-training alignment techniques based on human feedback~\cite{rafailov2023direct,liu2023statistical} or reward models~\cite{shao2024deepseekmath,yu2025dapo}, and intensively studied test-time computing techniques~\cite{wei2022chain}. Each component has played a vital role in enhancing model capabilities. These developments have endowed LLMs with powerful abilities in generation, comprehension, and reasoning, allowing them to exhibit strong generalization and emergent behaviors across a wide range of tasks.


Given the intrinsic similarity between autonomous driving motion generation and language generation tasks, several recent works~\cite{behaviorgpt,smart,zhang2025carplanner,drope,dauner2023parting,catk,unimm} have attempted to transfer modules from LLMs to this domain. In terms of tokenizer design, methods such as Trajeglish, MotionLM, and Kigras discretize agent motion trajectories into tokens, while BehaviorGPT~\cite{behaviorgpt}, SMART~\cite{smart}, and CarPlanner~\cite{zhang2025carplanner} further extend this approach by jointly tokenizing both scene context and multi-agent trajectories. For positional embedding, Drope~\cite{drope} draws on the concept of relative positional embedding from LLMs and proposes a strategy based on inter-agent spatial relationships, aiming to reduce the high GPU memory consumption associated with traditional coordinate system transformations such as SMART~\cite{smart}. Regarding training paradigms, many recent approaches~\cite{behaviorgpt,smart,zhang2025carplanner,drope} adopt autoregressive generation schemes inspired by LLMs and also benefit from the efficiency of the teacher-forcing method, generating future trajectories of agents in a closed-loop manner. At the post-training stage, similar to how LLMs align model behavior with human intent via preference-based fine-tuning, several motion generation methods—such as CarPlanner~\cite{zhang2025carplanner} and PDM~\cite{dauner2023parting}—introduce behavior optimization or safety-aware constraints to improve trajectory quality and reliability.

However, most existing studies only incorporate ideas from LLMs at the module level and lack a comprehensive and systematic analysis of the architectural and methodological differences between large language models and motion generation systems. To the best of our knowledge, this work is the first to conduct a system-level investigation, providing a thorough comparison between large language models (LLMs) and motion generation models across architectural design, training paradigms, and inference procedures. By examining the alignment and divergence between these two domains, we aim to offer insights and perspectives that may inform the development of future autonomous driving systems.

\subsection{Implementation Details}
\subsubsection{Sim Agents Metrics}
\label{subsubsec:sim_agents_metrics}

We adopt the evaluation metrics from the Waymo Open Sim Agents Challenge (WOSAC)~\cite{montali2023waymo} to assess the quality of our agent motion generation. The WOSAC metrics define realistic agents as those whose behavior aligns with the actual distribution of scenarios observed in real-world driving. To evaluate simulation performance, these metrics use a point-to-distribution divergence metric. Specifically, they compute the approximate negative log-likelihood (NLL) of real-world samples with respect to the distribution induced by the simulated agents~\cite{montali2023waymo}. The objective is to minimize this NLL, which is formally defined as:
\begin{equation}
\text{NLL}^* = -\frac{1}{|\mathcal{D}|} \sum_{i=1}^{|\mathcal{D}|} \log q^{\text{world}}(o_{\geq 1, i} \mid o_{<1, i})
\end{equation}

The model is required to generate 32 future trajectories for each agent in order to construct histograms for standardizing NLL computations. The evaluation metric then computes NLLs under the standardized histogram-based distribution. These NLLs are evaluated across 9 different metrics, covering kinematic metrics, object interaction metrics, and map-based metrics~\cite{montali2023waymo}. These metrics are combined into a single comprehensive score, referred to as the Realism Meta Metric, by computing a weighted average according to the following formula:
\begin{equation}
\mathcal{M}^K = \frac{1}{NM} \sum_{i=1}^{N} \sum_{j=1}^{M} w_j \cdot m_{i,j}^K, \qquad \sum_{j=1}^{M} w_j = 1
\end{equation}

where $N$ is the number of scenarios and $M = 9$ is the number of component metrics. The value $m_{i,j}$ represents the likelihood for the the $j$-th metric on the $i$-th example.

We next analyze the details of the nine evaluation metrics.
The kinematic metrics include linear speed, linear acceleration, angular speed, and angular acceleration magnitude. These metrics are used to assess the similarity between the kinematic properties of the model-generated trajectories and those of human driver trajectories.

Object interaction metrics evaluate the interaction behaviors between agents. These include measures such as the distance to the nearest object and time-to-collision, which assess the similarity to human driving behavior, as well as the collision likelihood metric, which evaluates how closely the model-generated collision frequency matches that observed in real-world trajectories. Given the critical importance of safety in driving, the collision likelihood metric is considered more significant and is assigned a higher weight. We observe that, in most cases, the collision likelihood score increases as the collision rate decreases. However, due to sensor noise or fluctuations in perception models, some ground-truth trajectories may inherently involve collisions. As a result, if the model generates only collision-free trajectories, it may be assigned a lower likelihood under the learned distribution, ultimately resulting in a lower evaluation score.

Map-based metrics include the distance to the road edge, which assesses the similarity between model-generated behavior and that of human drivers, and the off-road likelihood, which evaluates how closely the probability of map-edge collisions in generated trajectories matches those observed in human driving. Similar to the collision likelihood metric, offroad likelihood is assigned a higher weight due to its relevance to driving safety.
However, due to sensor noise and map annotation errors—such as incomplete labeling that marks vehicles entering parking areas as colliding with map boundaries—some ground-truth trajectories in the dataset may involve apparent collisions with the map. As a result, model-generated trajectories that avoid such regions may receive lower likelihood scores, ultimately leading to a lower overall evaluation score.

Additionally, metrics includes Average Displacement Error (ADE) and minimum Average Displacement Error (minADE). ADE averages Euclidean position errors over the forecast horizon, while minADE, given K samples, selects the per-scene minimum ADE across K. These two metrics are commonly used to measure how closely the generated results follow the vehicle’s actual trajectory.

\subsubsection{Baseline}
\label{appd:baseline}

Our network framework consists of two main components: a scene encoder and a GPT-based motion generator. The scene encoder, inspired by the scene encoding module in LTP~\cite{wang2022ltp}, compresses the map and agent history into instance-level tokens. The GPT-based motion generation module, inspired by MotionLM~\cite{motionlm}, adopts an auto-regressive approach to sequentially generate motion tokens in the Verlet format.

\subsubsubsection{\textbf{Scene Encoder}}
\label{baseline:scene_encoder}
Given an environmental scenario $E: \left \{ \mathcal{M}, \mathcal{A} \right \}$, where $\mathcal{M}=\left \{ m_{i} \right \}$ represents the lane segments and road edges in the scene and $\mathcal{A}=\left \{ a_{i} \right \}$ represents all agents' trajectories. The length of each map elements $m_{i}$ is cut to $l$m or less if it is shorter than $l$m and annotated as $m_{i} = \left \{ m_{i}^j, j\in [1, N] \right \} $. For lane segments, the value of $l$ is set to 10 meters, while for road edges, $l$ is set to 20 meters.  $N$ controls the resolution of the map. Each $m_i^j$ describes a section of the $\frac{l}{N}$-meters lane, defined as $[sx, sy, ex, ey, th, le, ty]$, where $(sx, sy)$ and $(ex, ey)$ are the start and end point of the vector, $th$ and $le$ represent the heading angle and the length of the vector, and $ty$ represents the type of map elements.
Similarly, for each agent trajectory $ a_i = \left \{ a_i^j, j \in [1, T] \right \} $, where $T$ represents the total steps of the trajectory, each $a_{i}^j$ describes a vector that makes up the trajectory, defined as $[sx, sy, ex, ey, th, le, ty, ts]$. The definitions of the first six dimensions are the same as $c_{i}^j$. $ty$ represents the type of agents, such as vehicles or cyclists. Further, $ts$ describes the absolute time relative to the current frame. In our implementation, map elements are uniformly sampled at 1-meter intervals. For the Sim Agents task, each agent’s motion history spans a duration of 1 second.

\begin{figure}[h]
    \centering
    \includegraphics[width=1.0\textwidth]{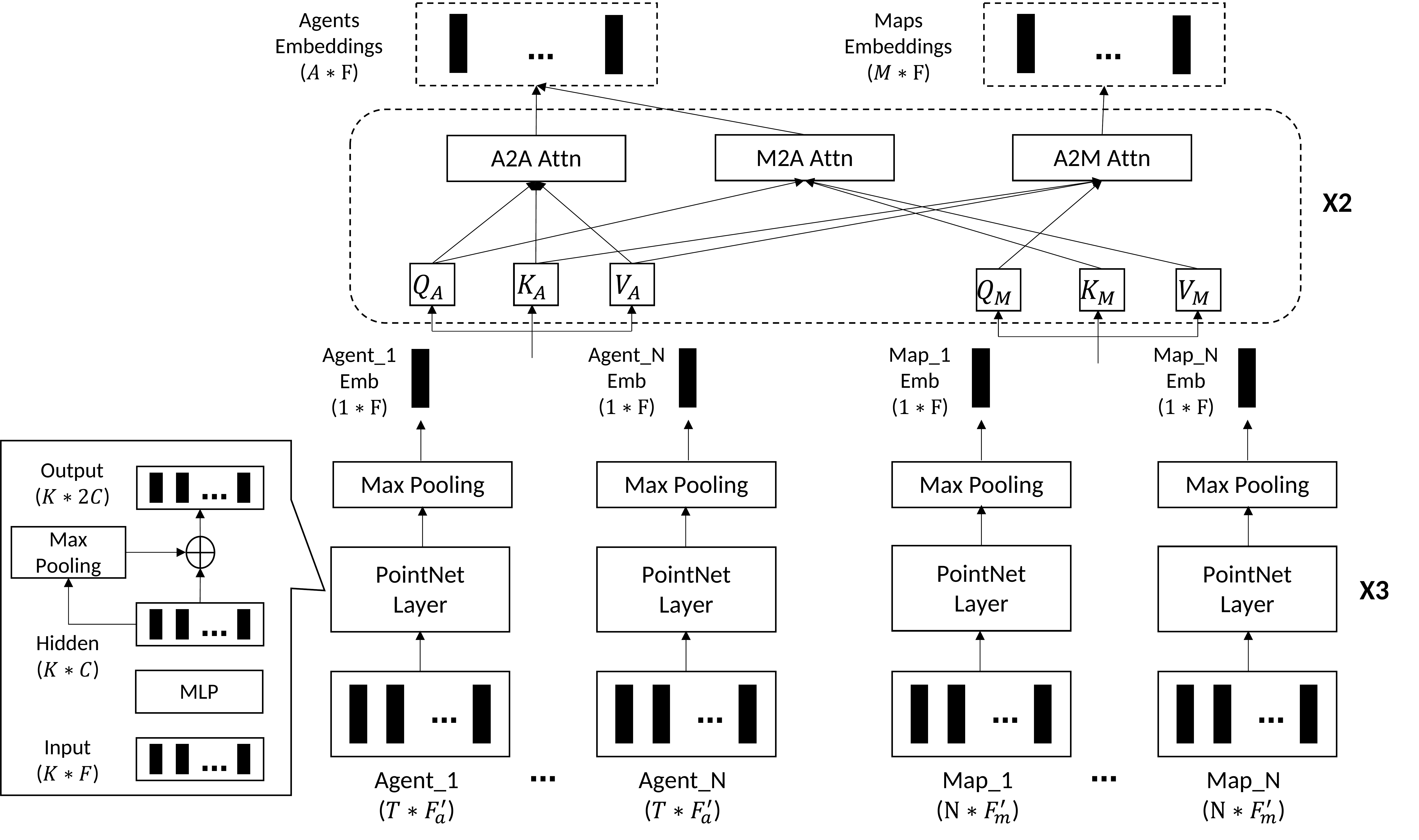}
    \caption{The architecture of the scene encoder.}
    \label{fig:scene_encoder}
\end{figure}

In the scene encoder, for each instance-level entity (i.e., lane segments, boundary lines, and agents), we first extract a fixed-dimensional feature vector given an input of shape $(K,F')$ using a PointNet-like module. Specifically, we adopt a three-layer ShapeNet, which progressively aggregates the input features via concatenation and max-pooling operations, ultimately producing a single feature vector of shape $(1, F)$. We then alternate between A2A self-attention, A2M cross-attention, and M2A cross-attention for two layers to enable information exchange among instances. Unlike prior works, we deliberately omit the M2M self-attention module due to its substantial memory overhead, which becomes especially problematic during reinforcement learning rollouts. Our empirical results show that removing M2M leads to approximately a $2\%$ drop in accuracy, but yields around a $30\%$ improvement in training and inference speed. We consider this a reasonable trade-off. The network architecture is illustrated in Figure \ref{fig:scene_encoder}. In our implementation, we set $F=128$.

\begin{figure}[h]
    \centering
    \includegraphics[width=1.0\textwidth]{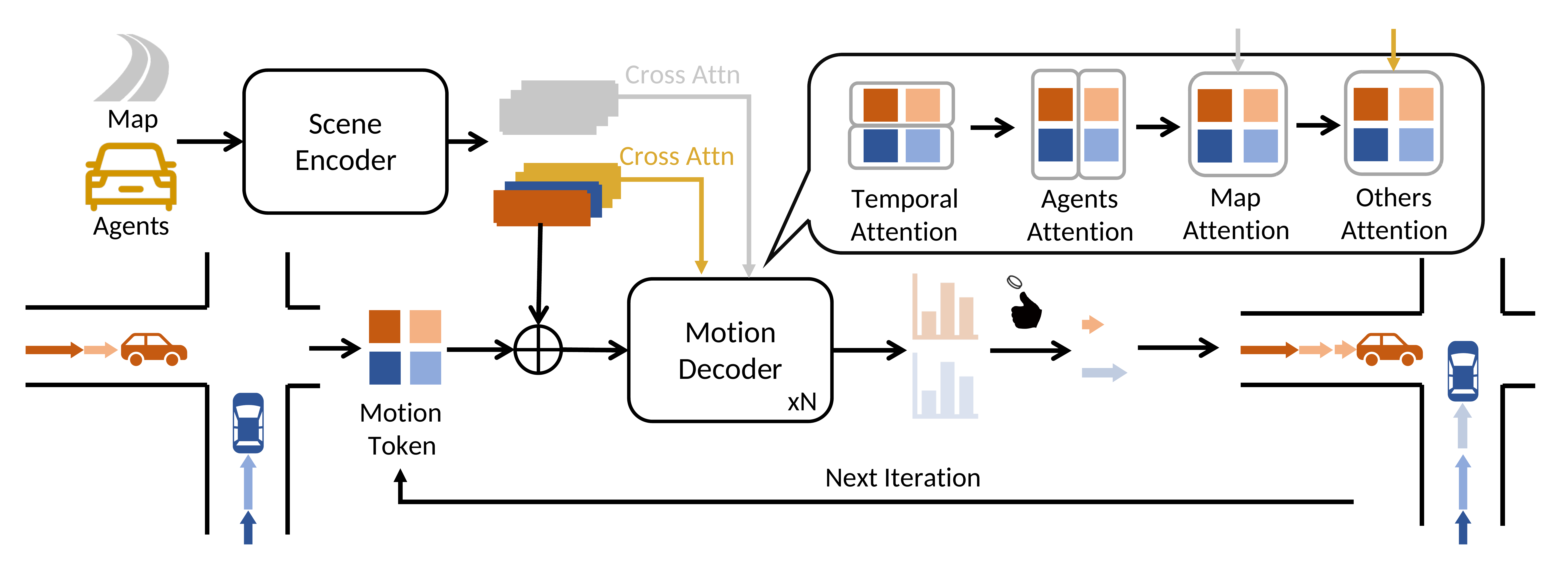}
    \caption{The architecture of the motion generator.}
    \label{fig:motion_generator}
\end{figure}

\subsubsubsection{\textbf{Motion Generator}}
\label{baseline:motion_generator}
Similar to MotionLM, we adopt an autoregressive generation scheme for multi-agent motion generation. However, MotionLM encodes the environment separately for each agent in their respective local coordinate systems. This design, while intuitive, suffers from the curse of dimensionality and thus primarily focuses on motion generation for only two agents. In contrast, we encode both the static environment and agent history in the scene-centric (self-driving car) coordinate frame, enabling a shared representation across all agents. As a result, we can jointly generate motion tokens for $A$ interest agents at each iteration. In the Sim Agents task, $A$ is set to 8 as the maximum.

To maintain awareness of the map and other agents throughout the autoregressive generation process, our GPT-based model incorporates several types of attention mechanisms during inference. Specifically, we perform: (1) self-attention over each agent’s motion tokens across different time steps; (2) self-attention between different agents at the same time step; (3) cross-attention over the static map context; and (4) cross-attention over non-predicted agents that are excluded from the GPT input during rollout. The detailed architecture and inference flow are illustrated in Fig \ref{fig:motion_generator}.

\subsubsection{Tokenizing}
\label{appendix:tokenizer_design}



\begin{figure}[h]
    \centering
    \includegraphics[width=0.8\textwidth]{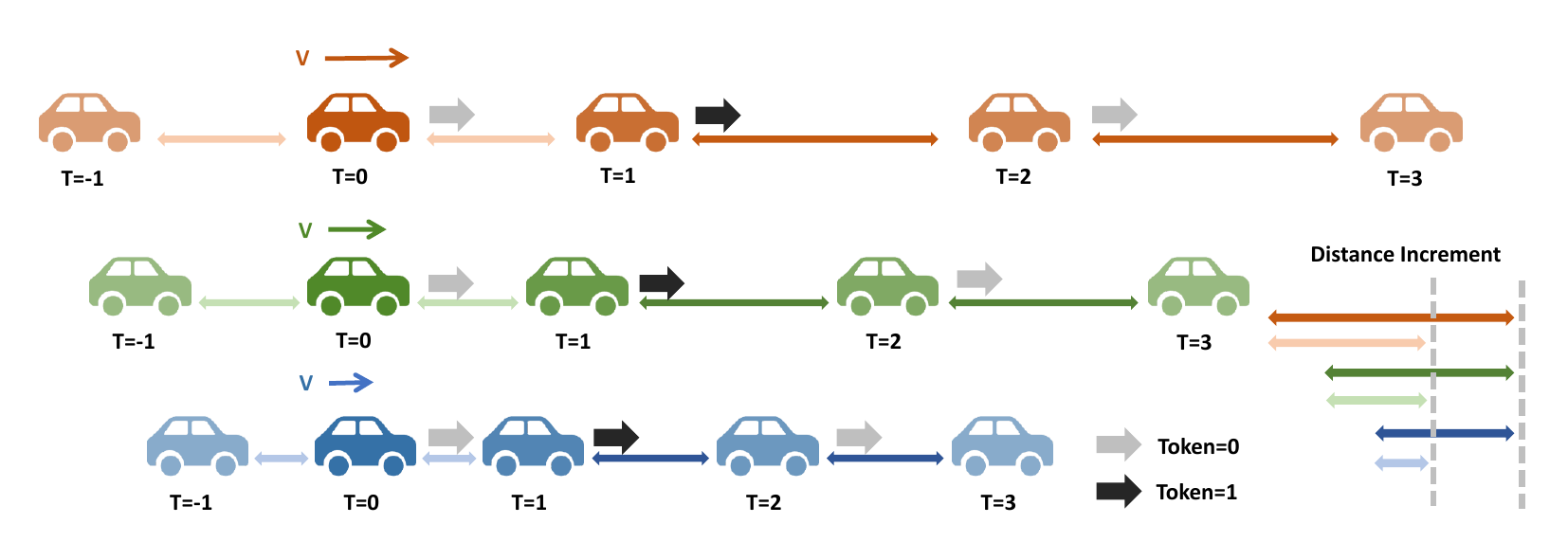}
    \caption{Influence of initial velocity on the travel distance induced by the same motion token sequence. A token sequence $[0, 1, 0]$ encoded by the Verlet Wrapper represents “maintain speed, accelerate, maintain speed.” The encoding is consistent across vehicles, but final positions vary depending on initial velocities. Since the same token value represents the same acceleration, the displacement increments of vehicles with different initial speeds are identical.}
    \label{fig:token_process}
\end{figure}

In trajectory tokenization, the process involves trajectory segmentation, discretization, and transformation through the Verlet Wrapper used in MotionLM~\cite{motionlm}. Owing to variations in the initial states, the same token may correspond to different physical motions depending on the initial state. Under this model-based encoding paradigm, the token can be interpreted as an implicit representation of acceleration. In this formulation, given the same acceleration, the resulting displacement increment remains consistent. This is illustrated in Figure~\ref{fig:token_process}.

The complete trajectory with duration $T = 9s$ (consisting of $1s$ historical trajectories and $8s$ future predicted trajectories) is segmented according to a specified time step. We select a segmentation time step of $\Delta t = 0.5$ seconds, resulting in a total of $18$ trajectory segments. 

In MotionLM~\cite{motionlm}, the tokenization process is based on the absolute changes in agent positions relative to a fixed reference point, usually the location of the self-driving car at the anchor time step. In our approach, we instead redefine the encoding within the local coordinate frame of each individual agent. We begin by transforming trajectories into an ego-centric coordinate frame, which provides a normalized reference for subsequent processing. In this frame, the position information, composed of X and Y components, is then processed separately. Each spatial dimension is discretized into $128$ bins within a length range of $[-18\text{m}, 18\text{m}]$. This discretization covers velocities ranging from $-36\text{m/s}$ to $36\text{m/s}$, encapsulating over $99\%$ observed vehicle trajectories~\cite{motionlm}.

After obtaining discrete trajectory information, we use the Verlet Wrapper to process these discrete positions. Using the discrete values from the initial $0.5s$ segment as a reference, we compute the differences between discrete values of consecutive trajectory segments, yielding 17 discrete trajectory differences. The discrete difference data spans an interval of $[-6, 6]$, resulting in a total of $13$ discrete values, which correspond to acceleration changes within $[-3.6 \text{m/s}^2, 3.6 \text{m/s}^2]$. After computing discrete differences separately for each spatial dimension, we combine the X and Y values to generate a total of $13 \times 13 = 169$ discrete trajectory encodings, which serve as tokens within the motion vocabulary.

\subsubsection{Positional embedding}
\label{appendix:positional_embedding}

In the \textbf{No PE} setting, our implementation adopts the following design choices: after the ShapeNet module in Figure~\ref{baseline:scene_encoder}, we add a learnable type embedding to each instance, separately for agents and lanes/road edges. Additionally, during the motion generator \ref{baseline:motion_generator}, we incorporate a learnable time-wise embedding into the sequential motion tokens to preserve temporal ordering. Apart from these, no additional positional embeddings are applied in any of the attention modules.

In the \textbf{Vanilla PE} setting, we retain the learnable components from the \textbf{No PE} baseline, including the type embeddings and time-wise embeddings. Additionally, we incorporate sinusoidal positional encodings to each token before it enters any attention module, following the most commonly used formulation~\cite{vaswani2017attention}:
\begin{equation}
\mathrm{PE}_{(pos, 2i)} = \sin\left(\frac{pos}{10000^{2i/d_{\text{model}}}}\right), \quad
\mathrm{PE}_{(pos, 2i+1)} = \cos\left(\frac{pos}{10000^{2i/d_{\text{model}}}}\right)
\label{eq:origin_pe}
\end{equation}

In the \textbf{DRoPE} setting, we follow the Directional Rotary Position Embedding~\cite{drope} method, in which positional and angular information are encoded using 2D-RoPE and Directional-RoPE, respectively. Specifically, for each instance $X_i$, its global positional coordinates $pos_i = (x,y)$ are incorporated into the input embedding via a rotation-based transformation as defined in Equation \eqref{eq:drope_pos}, in which $\theta=\{\theta_l\}_{l=1}^{d_k}$ corresponds to the angular component typically used in conventional positional embedding schemes. The global heading $\alpha_i$ is encoded using the method described in Equation \eqref{eq:drope_heading}. We adopt a head-by-head alternation strategy to inject positional and directional features into different attention heads. For a theoretical proof of the lossless encoding of relative position and orientation under this scheme, please refer to~\cite{drope}.
\begin{equation}
\hat{X}_i = R_{\text{pos}_i} X_i, \quad \text{where } R_{\text{pos}_i} =
\left(
\begin{array}{cccc}
\cos x\theta & -\sin x\theta & 0 & 0 \\
\sin x\theta & \cos x\theta & 0 & 0 \\
0 & 0 & \cos y\theta & -\sin y\theta \\
0 & 0 & \sin y\theta & \cos y\theta \\
\end{array}
\right)
\label{eq:drope_pos}
\end{equation}


\begin{equation}
\hat{X}_i = R_{\alpha_i}X_i,\\
\text{where } R_{\alpha_i} = \left( 
\begin{array}{cc} 
\cos \alpha_i & -\sin \alpha_i \\ 
\sin \alpha_i & \cos \alpha_i \\ 
\end{array}\right)
\label{eq:drope_heading}
\end{equation}

In the \textbf{Global-DRoPE} setting, while keeping all other configurations identical to \textbf{RoPE based positional embedding}, we encode all the raw input data from ShapeNet in the SDC coordinate system, meaning that different lane segments and agents are no longer encoded in their own local coordinate systems. This can increase the diversity of “words” within the scene.

\subsubsection{Pre-training}
\label{appd:pre-training}

Motion generation task can be formulated as a next-token prediction problem, where the goal is to estimate the probability distribution of the next token conditioned on the preceding ones. The full trajectory over the prediction horizon is then generated iteratively through an autoregressive process. 

In the WOSAC benchmark, models are tasked with predicting future trajectories over an 8-second horizon, given 1 second of past trajectory history. Typically, the anchor time for encoding contextual information---including the agent’s motion history and map topology---is set to \( t = 1 \) second. However, trajectory generation depends not only on token sequences but also critically on rich contextual cues. Yet, relying solely on a single anchor time for autoregressive training is insufficient to fully leverage such information.

To address this limitation, we propose a data augmentation strategy tailored for autoregressive models. In addition to training at the current anchor time \( T_\mathit{anchor} = 1s \), we introduce multiple generation anchors spaced at intervals of \( t_{\mathit{interval}} \). At each anchor, we construct token sequences of varying lengths \([T_{anchor} + t_{\mathit{interval}},\  T_\mathit{anchor} + 2*t_{\mathit{interval}},\  \ldots,\  T_{anchor} + K*t_{\mathit{interval}}\)] along with their corresponding contextual information for autoregressive training. The full training procedure is detailed in Algorithm~\ref{alg:training}. In the actual training process, we set $K=8$ to achieve an eightfold data augmentation.
\begin{algorithm}[H]
\caption{Pre-train and fine-tune an autoregressive motion predictor}
\label{alg:training}
\begin{algorithmic}[1]
\Require Large-scale driving dataset $\mathcal{D}$
\Ensure A Sim Agents policy $\pi_\theta$
\State Initialize model parameters $\theta$ and model $\pi_\theta$
\For{pre-training epoch $j = 1, \dots$}
    \State Retrieve training samples $\{\mathbf{s}_{\mathit{history}}, \mathbf{s}_{\mathit{GT}}, \mathcal{M}\}$ from $\mathcal{D}$
    \For{each anchor time $T' \in \{T_{anchor} + k \cdot t_{\mathit{interval}} \mid k = 0, 1, 2, \dots, K-1\}$}
        \State Construct token sequence and context at anchor $T'$
        \State Construct target actions $\mathbf{a}_{\mathit{GT}} = \{a_{t,i}^{\mathit{GT}}, \forall t, i\}$ and observation $o_{t,i}$ with GT actions
        \State Run the model with the observation and get the predicted actions $\{a_{t,i}, \forall t, i\}$
        \State Update $\pi_\theta$ via $\mathcal{L}(\theta) = - \mathbb{E}_{(x_{1:T}, c) \sim \mathcal{D}} \left[ \sum_{t=1}^{T} \log p_\theta(x_t \mid x_{1:t-1}, c) \right]$
    \EndFor
\EndFor
\end{algorithmic}
\end{algorithm}

\begin{figure}[htbp]
    \centering 
    \vspace{-0.2cm}
    \subfloat[Effect of Data Volume on Train Loss]{
    \includegraphics[width=0.43\linewidth]{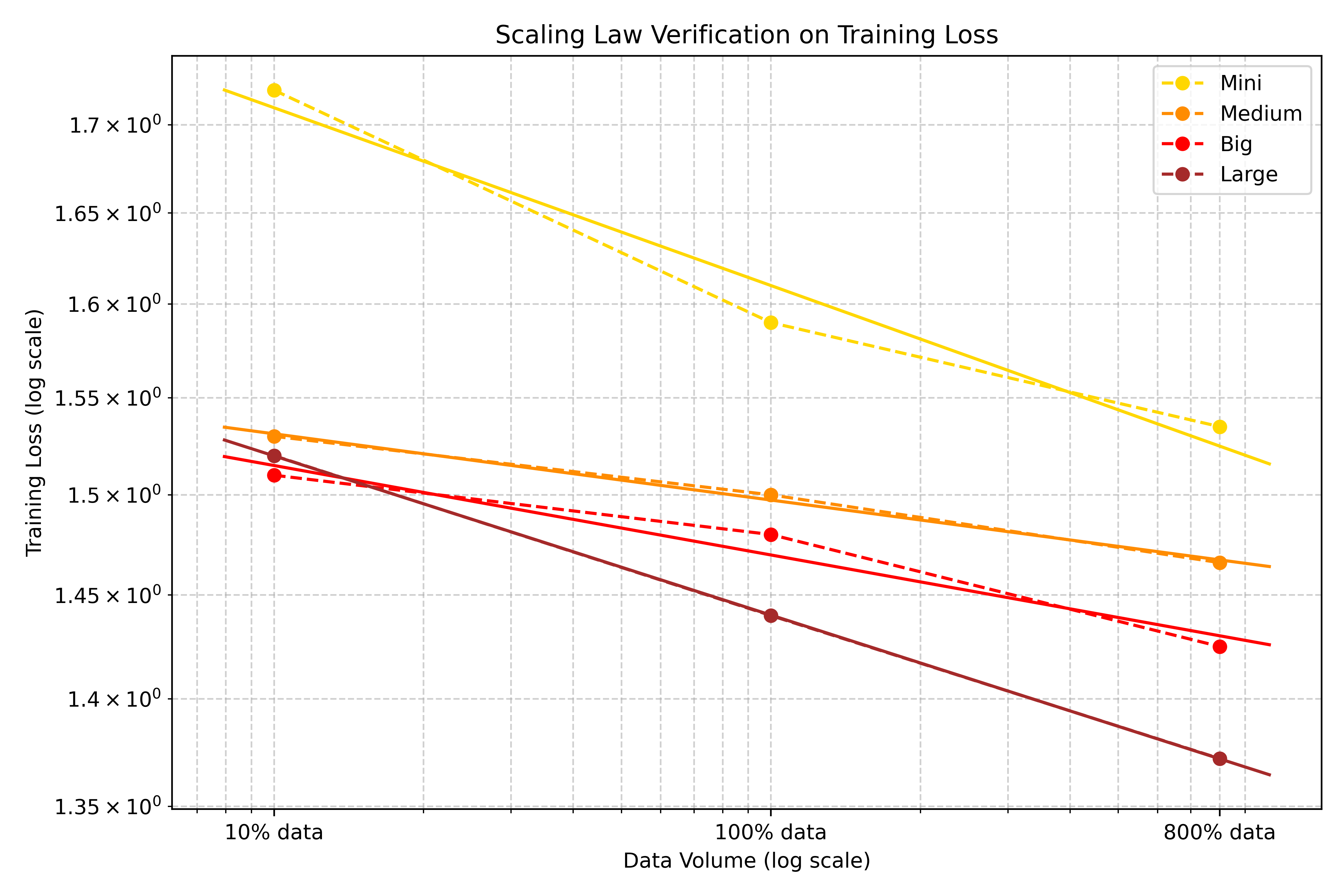}\label{1}}
    \subfloat[Effect of Data Volume on Val Loss]{
    \includegraphics[width=0.43\linewidth]{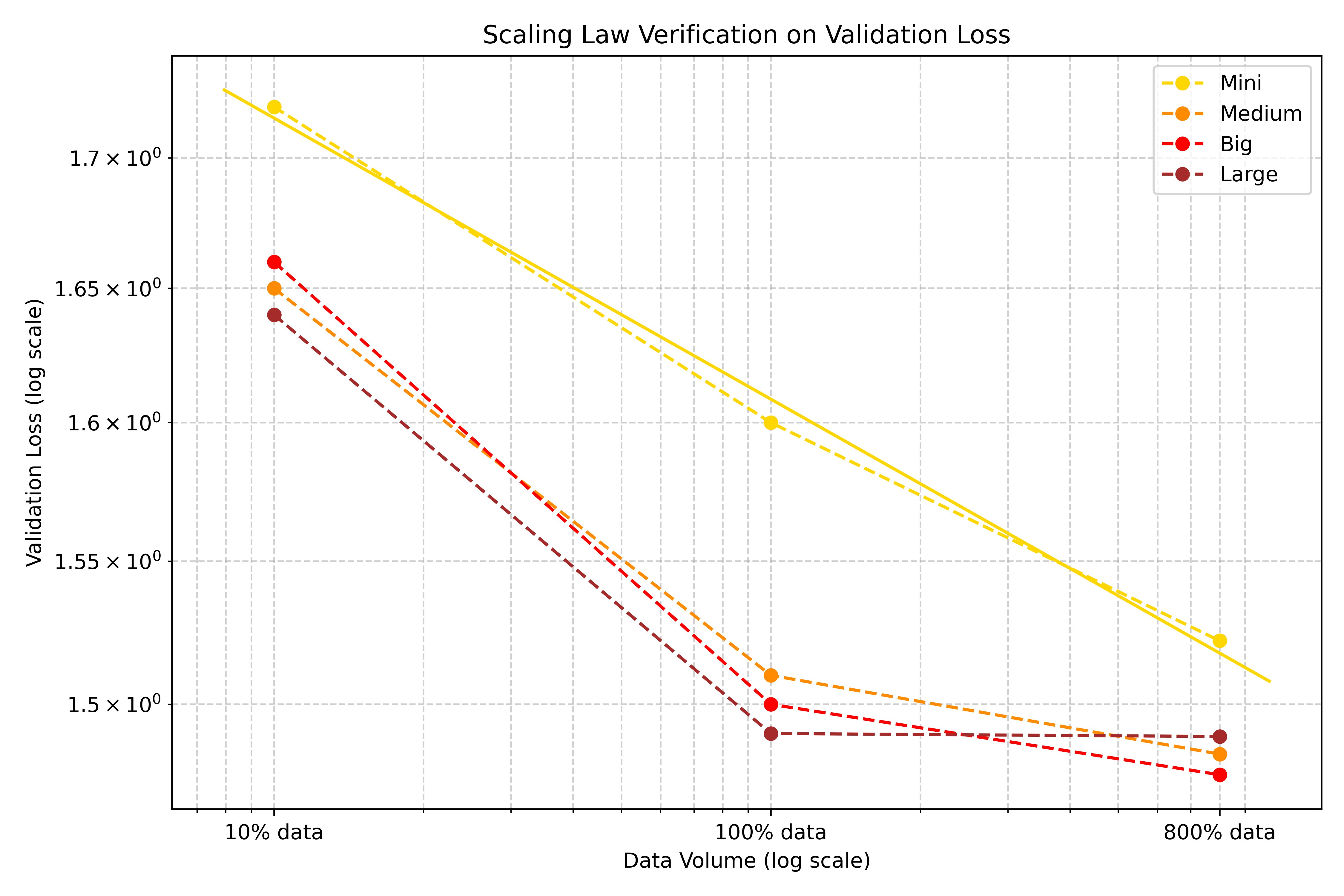}\label{2}}\\
    \subfloat[Effect of Model Parameters on Train Loss]{
    \includegraphics[width=0.43\linewidth]{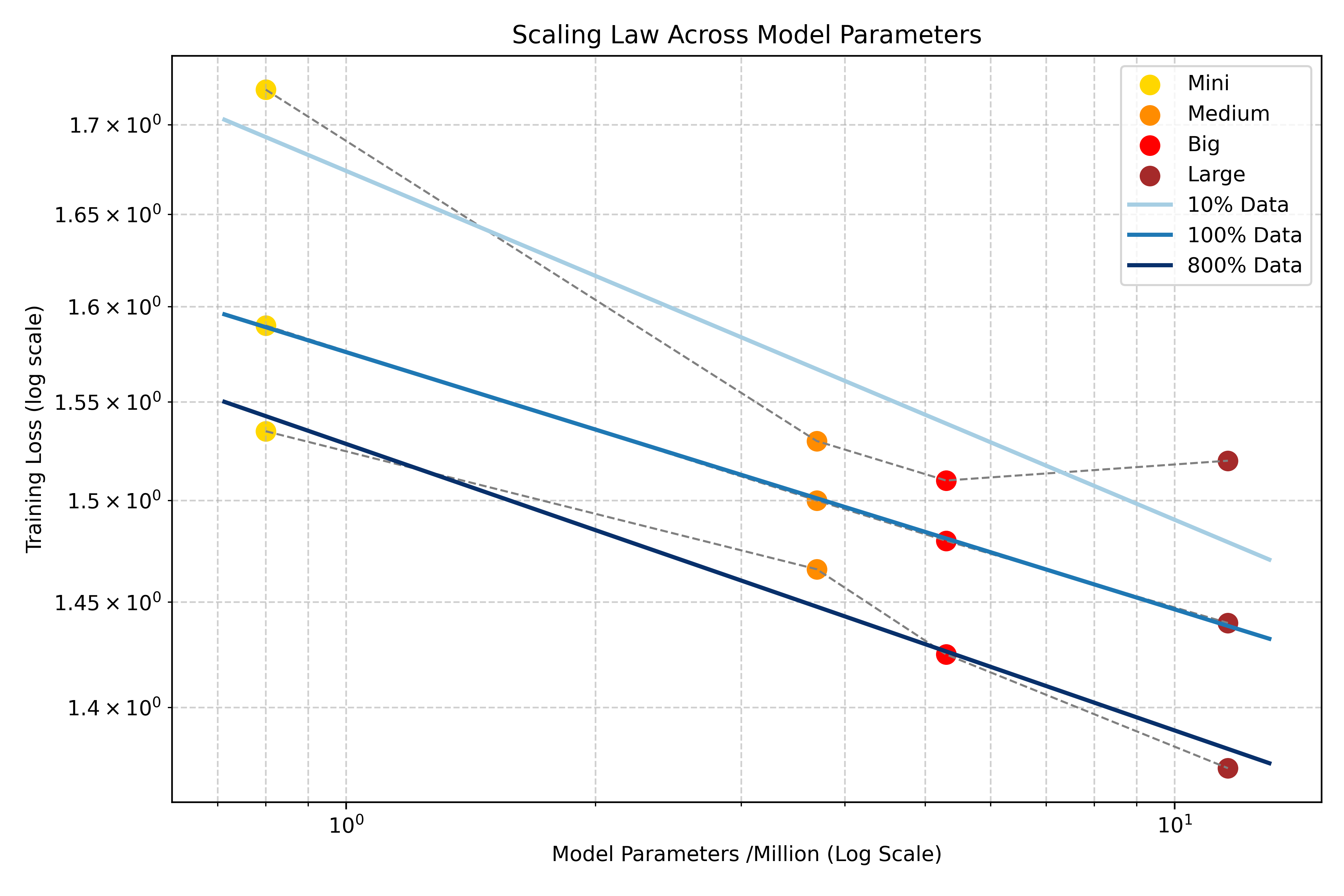}\label{3}}
    \subfloat[Effect of Model Parameters on Val Loss]{
    \includegraphics[width=0.43\linewidth]{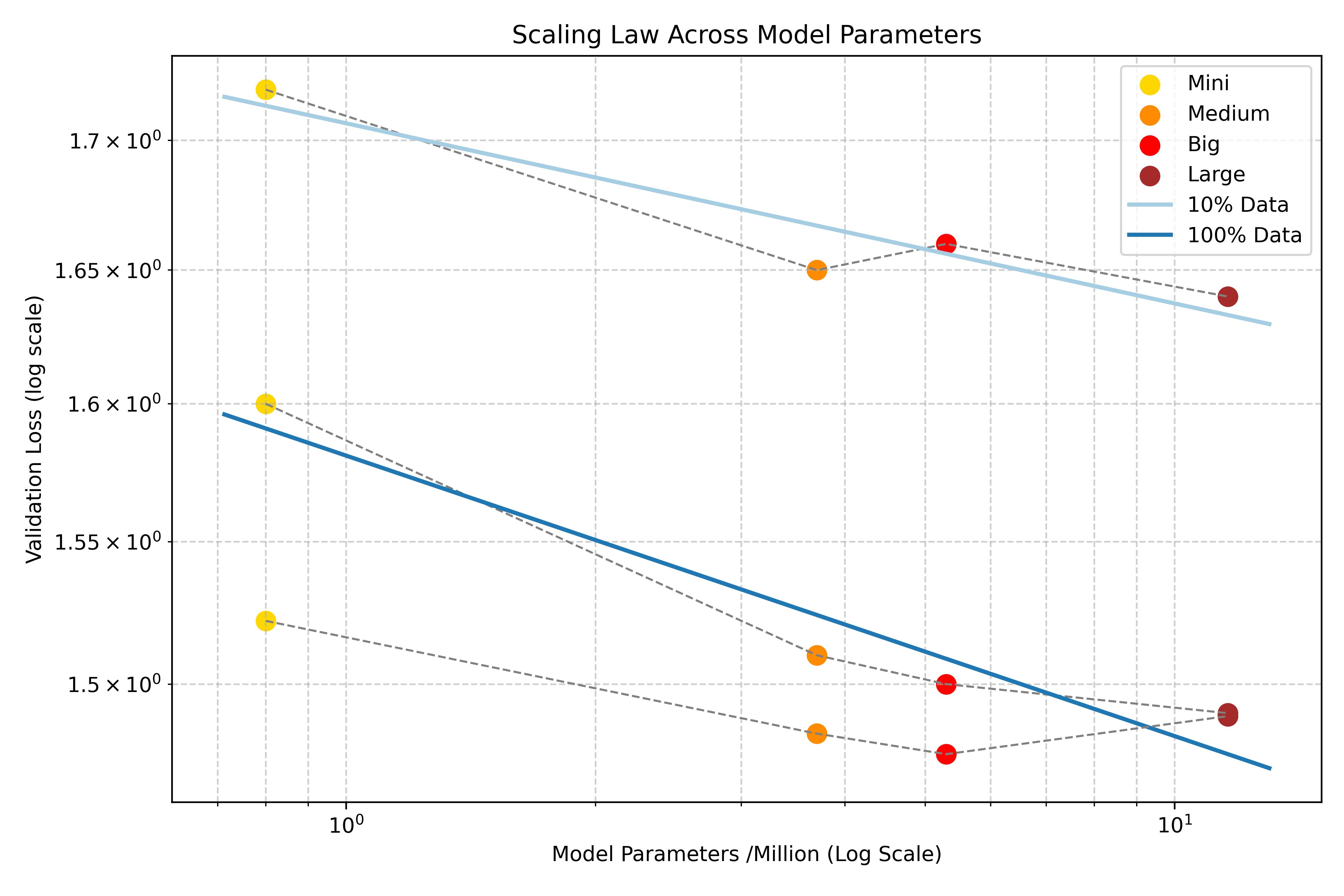}\label{4}}
    \caption{Scaling Law of Data Volume and Model Parameters\label{fig:scaling_law_vis}}
\end{figure}

We analyze the Scaling Law with respect to both training data volume and model parameter size as shown in Figure~\ref{fig:scaling_law_vis}. 

Figures (a) and (b) illustrate the training and validation losses, respectively, across various data scales for different model sizes. For all models, the loss decreases rapidly as the training data volume increases, demonstrating the expected trend under the scaling law. However, in Figure (b), the performance trend from 100\% to 800\% of the dataset exhibits a slight deviation from the ideal scaling behavior. This discrepancy arises because the augmented data used to simulate an 800\% dataset introduces high redundancy and lacks the diversity of a truly expanded dataset, thereby weakening the expected scaling effect.

Furthermore, even when trained on the largest available dataset (800\%), the Large model still exhibits signs of overfitting. This suggests that performance is constrained more by the limited diversity and size of the training data than by the model's capacity.

The scaling law holds when sufficient training data is provided: model performance improves consistently as the number of parameters increases. However, for the Large model, the 800\% data volume remains relatively small, making it prone to overfitting despite its higher capacity.

\subsubsection{Post-training}
\label{appendix:post_training}
Prior to post-training, we obtain a base model $\pi_{pre}$ through pre-training. For a given target agent, $\pi_{pre}$ can generate multiple future trajectories in parallel via next-token prediction, which can be written as ${\{\zeta_1, \zeta_2, ...,\zeta_N\}}$, where $\zeta_i=(s_0^i,a_0^i,s_1^i,a_1^i, ..,s_t^i,a_t^i)$. The primary objective of post-training is to further optimize the model using the outcomes $\mathbf{\{\zeta\}}$ of closed-loop generation.

In the \textbf{Supervised Fine-Tuning (SFT)} setting, we performed rollout simulations on the training set and identified cases involving collisions. These were selected to construct a small-scale fine-tuning dataset. On this dataset, we continued to apply the teacher-forcing strategy used during pre-training, with the only modification being a reduced learning rate set to $3e^{-5}$.

Motivated by the need for efficient sampling during reinforcement learning, we first constructed a lightweight GPU-based simulation environment for collision checking, similar in spirit to Waymax~\cite{waymax}. Within this environment, the rollout agents interact with both dynamic and static traffic participants, as well as map sidelines, for collision detection and reward feedback. Specifically, if an action leads to a collision with an obstacle, a penalty of $r_{\text{collision}} = -1$ is assigned; if it results in a collision with the road boundary, the penalty is $r_\text{offroad}=-1$; otherwise, the reward is set to $0$.
\begin{figure}[htbp]
    \centering
    \vspace{-3mm}
    \includegraphics[width=0.6\textwidth]{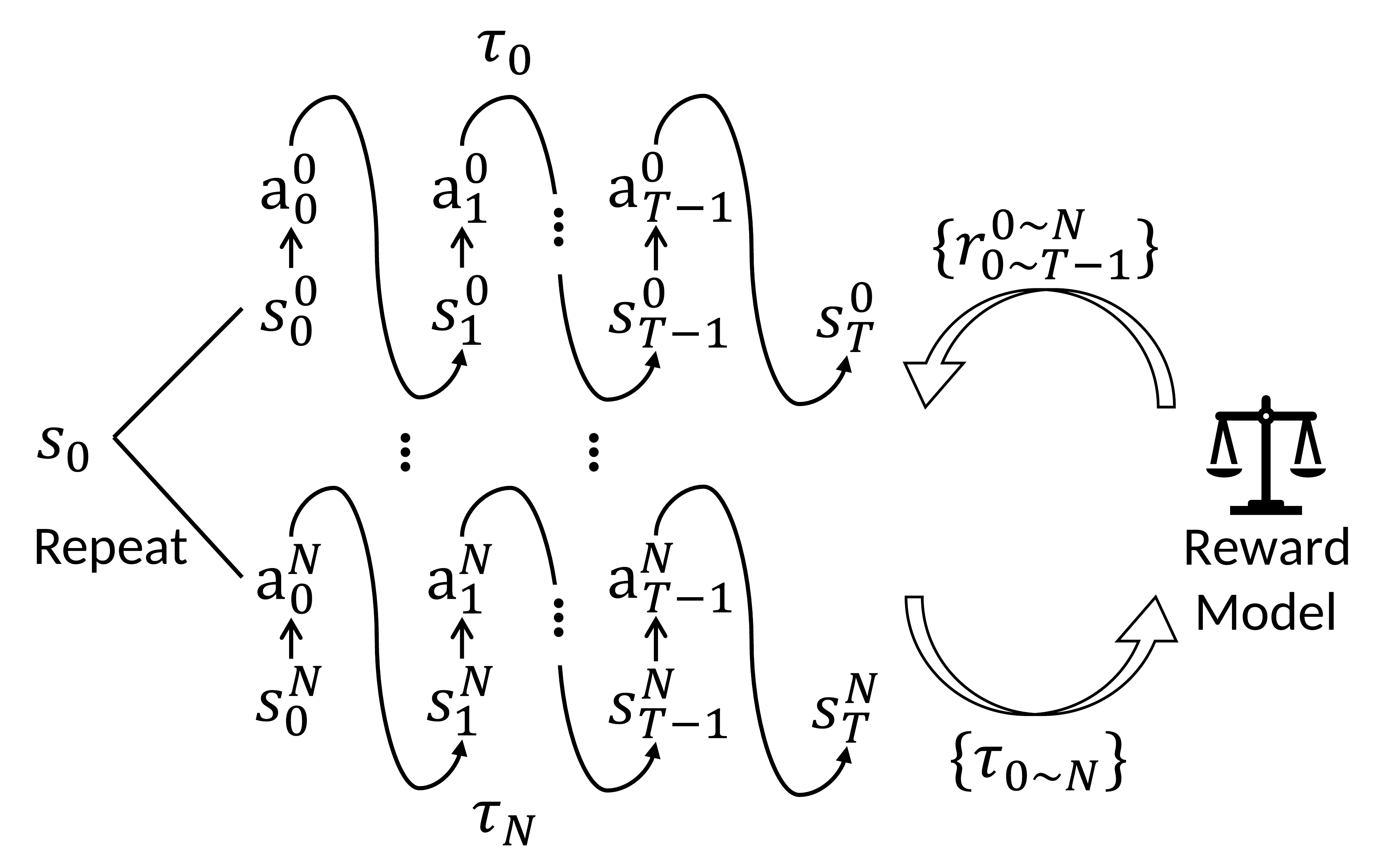}
    \vspace{3mm}
    \caption{In our reinforcement learning framework, each transition from $s_t^i$ to $s_{t+1}^i$ corresponds to one step inside the motion generator \ref{fig:motion_generator}. Moreover, the rollouts of $N$ trajectories can be executed in parallel. The reward model evaluates all $N$ rollouts at once and returns the corresponding step-wise rewards, following a structure similar to the reinforcement learning setup in LLMs.}
    \label{fig:post_training_framework}
\end{figure}

\vspace{-5mm}

Our reinforcement learning framework is illustrated in Figure~\ref{fig:post_training_framework}. Unlike traditional gym-like paradigms, our RL process more closely resembles reinforcement learning paradigms used in large language models. Each motion token generated by the motion generator is treated as an \textbf{action}, while each recursively fed query token is regarded as a \textbf{state}. The \textbf{reward} signals are provided by our custom lightweight GPU-based collision detector, which evaluates the safety of generated trajectories in real time.

In this framework, parallel generation of $N$ trajectories is straightforward: we simply replicate the input tokens $N$ times and feed them into the motion generator for parallel rollout. This setup leverages the autoregressive nature of the model while maintaining computational efficiency. Importantly, parallelization is applied only at the initial state, after which $N$ independent rollout trees are formed—each representing a separate exploration branch. Reinforcement learning fine-tuning is then applied on these parallel branches to optimize for safety and realism, while preserving the diversity of generated behaviors.

In the \textbf{REINFORCE} setting, we compute the return for each state-action pair by performing a discounted sum of the rewards along each closed-loop trajectory. These returns are then used to perform gradient ascent with respect to the objective defined in Equation \eqref{eq:reinforce}.
\begin{equation}
\mathcal{L}_{\text{RL}}(\theta) = -\mathbb{E}_{\pi_\theta} \left[ \sum_{t=1}^T R_t \cdot \log \pi_\theta(a_t \mid s_t) \right]
\label{eq:reinforce}
\end{equation}

In the \textbf{A2C} setting, we additionally train a critic network to serve as a baseline for estimating the advantage. The critic shares the same scene encoder with the actor, while the remaining components follow the same configuration as in the REINFORCE setup. The resulting objective function is given in Equation \eqref{eq:a2c}.
\begin{equation}
\mathcal{L}_{\text{RL}}(\theta) = -\mathbb{E}_{\pi_\theta} \left[ \sum_{t=1}^T (R_t - V_\theta(s_t)) \cdot \log \pi_\theta(a_t \mid s_t) \right]
\label{eq:a2c}
\end{equation}


In the \textbf{GRPO} setting, the critic network is no longer required. Instead, for all rollouts originating from the same initial state $s_0$, we perform within-group normalization of the return values at each time step: $\hat{A}_{i, t} = \widetilde{r}_i = \frac{r_i - \mathrm{mean}(\mathbf{r})}{\mathrm{std}(\mathbf{r})}$. This approach eliminates the need for explicit value estimation while still capturing relative return differences across parallel trajectories. The resulting optimization objective of GRPO is formulated in Equation~\eqref{eq:grpo_final}.

Compared to REINFORCE and A2C, the GRPO objective additionally incorporates a KL divergence term to constrain the policy from deviating excessively from the pre-trained model, thereby preserving human-likeness, and includes an entropy bonus to encourage greater diversity in the generated behaviors. It is worth noting that in this setting, we compute the KL divergence for each action across $R$ rollouts, $A$ agents, and $T$ time steps. To incorporate this into the final loss function, we aggregate the KL divergences by taking the mean over the rollout and agent dimensions $(R, A)$, and the maximum over the temporal dimension $(T)$, as defined in Equation~\eqref{eq:kl-loss-mean-max} and Equation~\eqref{eq:kl-loss}. This design ensures that the KL divergence signal is not overly smoothed by temporal averaging, allowing the loss to remain sensitive to large discrepancies at any time step, which is particularly important for penalizing sudden unsafe behaviors.

Additionally, we introduce an entropy term into the policy during training to ensure that the strategy maintains a certain level of exploration. The entropy bonus is formulated in Equation~\eqref{eq:entropy} and encourages non-deterministic behavior, which helps prevent premature convergence to suboptimal solutions.
In our implementation, we set $\lambda_{\mathrm{KL}} = 0.8$ and $\lambda_H = 0.01$ to balance constraint and diversity during training.


\begin{equation}
\mathcal{L}_{\text{RL}}(\theta) = -\mathbb{E}_{\pi_\theta} \left[ \sum_{t=1}^T \hat{A}_{t} \cdot \log \pi_\theta(a_t \mid s_t) \right]
+ \lambda_{\mathrm{KL}} \; \mathcal{L}_{KL}
- \lambda_{H} \; \mathcal{H}\Big(\pi_\theta(a_t \mid s_t)\Big) 
\label{eq:grpo_final}
\end{equation}

\begin{equation}
\mathcal{L}_{KL} = \frac{1}{|R||A|} \sum_{r=1}^{|R|} \sum_{i=1}^{|A|} \max_{t} \, KL_{r,i,t}
\label{eq:kl-loss-mean-max}
\end{equation}

\begin{equation}
    KL_{r,i,t} = \pi_{\theta}(a_{r,i,t}|s_{r,i,t}) \log \frac{\pi_{\theta}(a_{r,i,t}|s_{r,i,t})}{\pi_{ref}(a_{r,i,t}|s_{r,i,t})}
    \label{eq:kl-loss}
\end{equation}

\begin{equation}
\mathcal{H}\Big(\pi_\theta(a_t \mid s_t)\Big)
= - \sum_{a_t} \pi_\theta(a_t \mid s_t) \log \pi_\theta(a_t \mid s_t)
\label{eq:entropy}
\end{equation}

\clearpage
\subsubsection{Test-time Computing}
The detailed procedure of our test-time computing strategy is outlined in Algorithm \ref{alg:test_time_computing}. Here, we reuse the parallel rollout generation method shown in Figure \ref{fig:post_training_framework} to produce diverse rollouts. Specifically, during clustering, we concatenate the final poses of all interest agents in each rollout to form a feature vector representing that rollout. We then apply K-Medoids clustering on these feature vectors, and the resulting cluster centers are selected as the final outputs. The detailed procedure is presented in Algorithm \ref{alg:cluster}.
\begin{algorithm}[htbp]
    \centering
    \caption{Test-Time Computing Procedure}
    \label{alg:test_time_computing}
    \begin{algorithmic}[1]
        \Require Trained model $\pi_\theta$, initial environment context $s_e$, initial query agents $s_q$, number of output rollouts $N$, number of batch rollouts $B$, bool $use\_search$, bool $use\_cluster$
        \Ensure Final selected rollouts $\tau^*$
        \State \textbf{Initialize} candidate set $\mathcal{T} \leftarrow \emptyset$
        \If {$use\_cluster$}
            \State set $N_{out} = K \times N$
        \Else 
            \State set $N_{out} = N$
        \EndIf
        \While {len($\mathcal{T}$) $<$ $N_{out}$}
            \State Repeat $s_0$ $B$ times $\rightarrow$ $s_b = \{s_0, s_0, ..., s_0\}$
            \State Sample trajectory batch $\tau_b \sim \pi_\theta(s_b)$
            \If {$use\_search$}
                \State Evaluate $\tau_b$ for collisions and boundary violations
                \For{each feasible $\tau_i \in \tau_b$}
                    \State Add $\tau_i$ to $\mathcal{T}$
                \EndFor
            \Else
                \State Add all $\tau_b$ to $\mathcal{T}$
            \EndIf
        \EndWhile
        \If {$use\_cluster$}
            \State $\tau^* = cluster(\mathcal{T})$
        \Else 
            \State $\tau^* = \mathcal{T}$
        \EndIf
        \State \Return $\tau^*$
    \end{algorithmic}
\end{algorithm}
\begin{algorithm}[htbp]
    \centering
    \caption{Rollouts Clustering Procedure}
    \label{alg:cluster}
    \begin{algorithmic}[1]
        \Require Batch rollouts $\tau^N$, Number of interested agents $K$
        \Ensure Final selected rollouts $\tau^*$
        \For {$n = 1$ to $N$}
            \State Initialize $f_n$
            \For {$a = 1$ to $K$}
                \State $f_n.\text{append}(Agent_a.\text{end\_pose}())$
            \EndFor
        \EndFor
        \State $cluster\_center = \text{K-Medoids}(f_N)$
        \State $\tau^* = \tau^N[cluster\_center]$
        \State \Return $\tau^*$
    \end{algorithmic}
\end{algorithm}

\clearpage
\subsubsection{Benchmark Configuration}
\label{sec:benchmark_config}

\begin{table}[htbp]
\centering
\caption{Detailed Configuration Parameters}
\label{tab:config_detail}
\renewcommand{\arraystretch}{1.3}
\resizebox{0.90\textwidth}{!}{
\begin{tabular}{p{2.5cm} p{5.5cm} p{2.5cm}}
\toprule
\textbf{Module} & \textbf{Parameter} & \textbf{Value} \\
\midrule
\multirow{5}{*}{Tokenizing} 
  & Type & Verlet-Agent \\
  & Step Frequency & 2Hz \\
  & Bins Per Coordinate & 128 \\
  & \makecell[l]{Token Number Per Coordinate} & 13 \\
  & Vocabulary Size & 169 \\
\midrule
\multirow{2}{*}{\makecell{Positional \\ Embedding}}
  & Embedding method & Global-DRoPE \\
  & \makecell[l]{Posi and Direc Combined Method} & Head by head \\
\midrule
\multirow{6}{*}{Pre-training}
  & Hidden Size $d_{\text{model}}$ & 128 \\
  & Scene Encoder Layer Number & 2 \\
  & Autoregressive Decoder Layer & 4 \\
  & Model Parameters & 5.3M \\
  & Pre-training Epoch & 40 \\
  & Pre-training Learning Rate & 1e-3 to 1e-5 \\
\midrule
\multirow{9}{*}{Post-training}
  & Sample Temperature & 1.0 \\
  & RL Method & GRPO \\
  & Discount Factor $\gamma$ & 0.5 \\
  & Coll. Weight Balance Agent and Map & 0.5 \\
  & KL Diverse Weight $\lambda_{\mathrm{KL}}$ & 0.8 \\
  & Entropy Weight $\lambda_{H}$ & 1e-2 \\
  & Fine-tune Learning Rate & 1e-5 \\
  & Post-training Epoch & 1 \\
\midrule
\multirow{3}{*}{\makecell{Testing-time \\ Computing}}
  & Sample Temperature & 1.0 \\
  & More Rollouts Number & 1024 \\
  & Cluster Method & K-Medoids \\
\bottomrule
\end{tabular}
}
\end{table}


\subsection{Additional Qualitative Results}

\subsubsection{Positional Embedding}
\begin{figure}[htbp]
    \centering
    \includegraphics[width=1.0\textwidth]{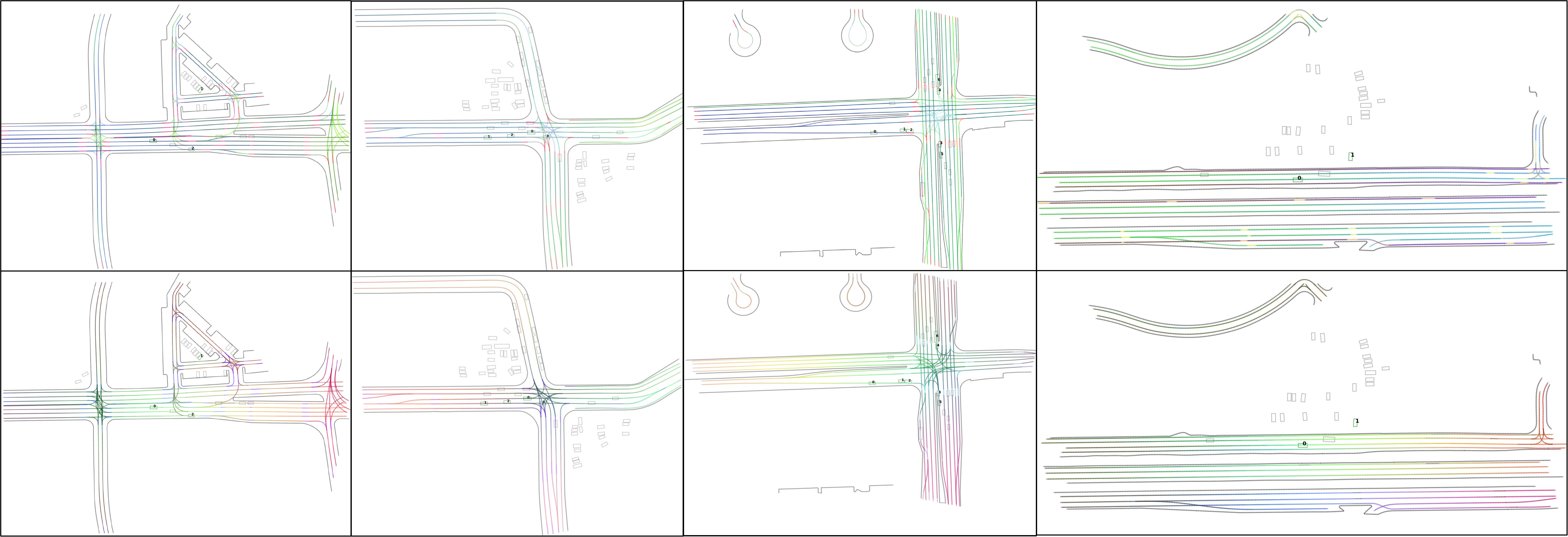}
    \caption{Visualization of lane segment features under DRoPE using inputs from local (top row) and global (bottom row) coordinates. Each 128-dimensional lane feature is projected into a 3D RGB space using PCA, such that color differences reflect the feature similarity across lane segments.}
    \label{fig:drope_lane_feat}
\end{figure}

We qualitatively compare the lane segment features produced by the scene encoder under the original DRoPE~\cite{drope} and Global-DRoPE settings, as illustrated in Figure~\ref{fig:drope_lane_feat}. As shown, lane features under the original DRoPE setting exhibit high redundancy—significant variations appear only when local attributes such as lane type, length, or curvature change. Consequently, even though the attention mechanism under DRoPE attends to lane segments at different distances, the corresponding \textbf{value embeddings} ($\mathbf{V}$) remain overly similar \footnote{In standard RoPE, only the query ($\mathbf{Q}$) and key ($\mathbf{K}$) vectors are subjected to rotational transformation, while the value ($\mathbf{V}$) vectors remain unaltered.}, which limits the model’s expressiveness. In contrast, globally encoded lane features vary more smoothly and effectively capture functional properties such as absolute position, orientation, and connectivity. Empirically, the combination of global encoding with DRoPE also yields superior performance. This highlights a key distinction between motion generation in autonomous driving and natural language processing: in NLP, tokens (i.e., words) inherently carry diverse semantics, whereas in motion generation, lane segments—serving as fundamental “tokens”—can become semantically indistinguishable when encoded in local frames, potentially leading to severe ambiguity.

\begin{figure}[htbp]
    \centering
    \includegraphics[width=1.0\textwidth]{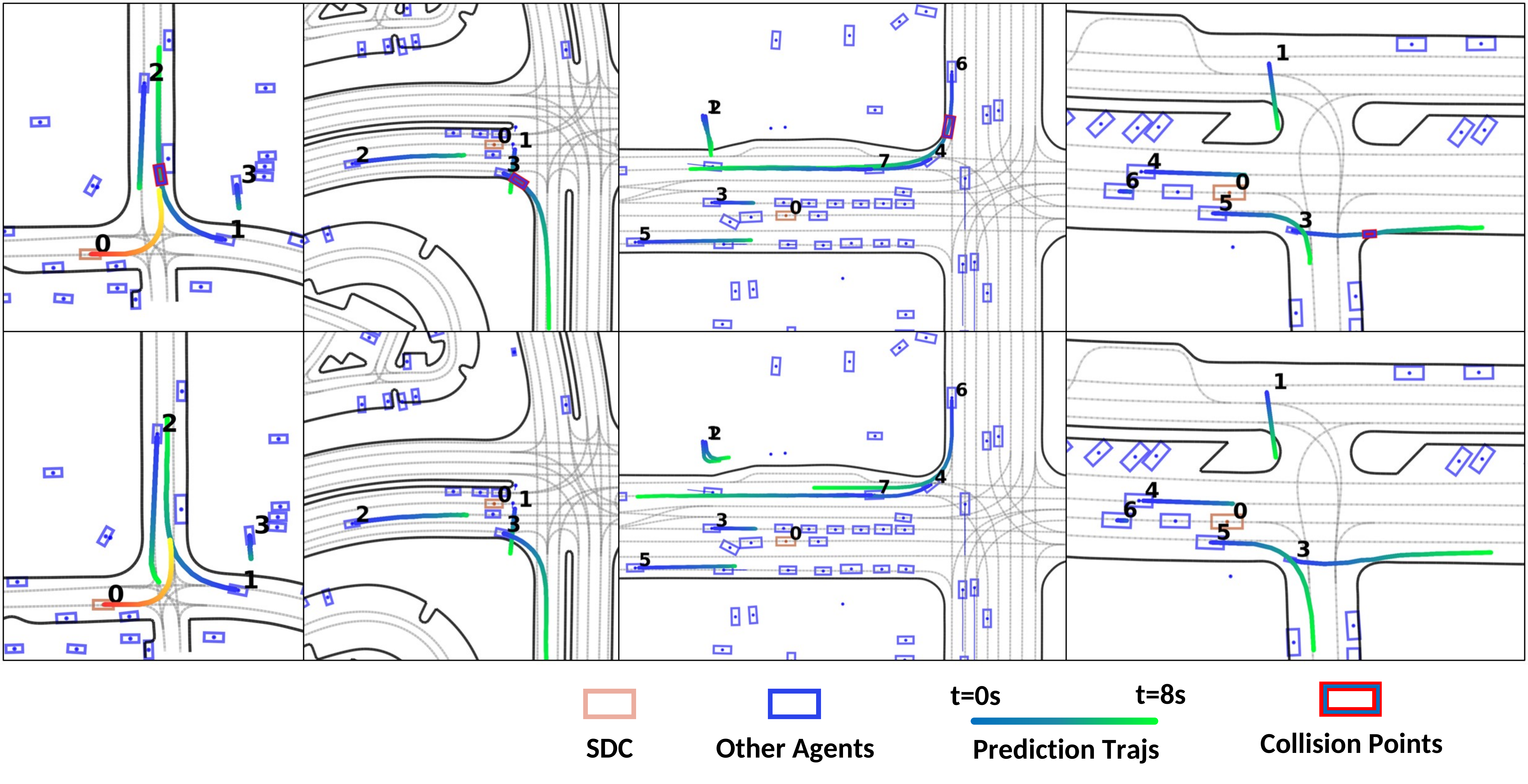}
    \caption{Under the DRoPE setting, we compare motion generation results using local (top row) and global (bottom row) coordinate encodings. It can be observed that in scenarios requiring precise understanding of lane and boundary geometry, the global coordinate input leads to more accurate trajectory generation, ensuring safer and collision-free motion.}
    \label{fig:drope_collision_comprision}
\end{figure}

\subsubsection{Pre-training}
\begin{figure}[htbp]
    \centering
    \includegraphics[width=1.0\textwidth]{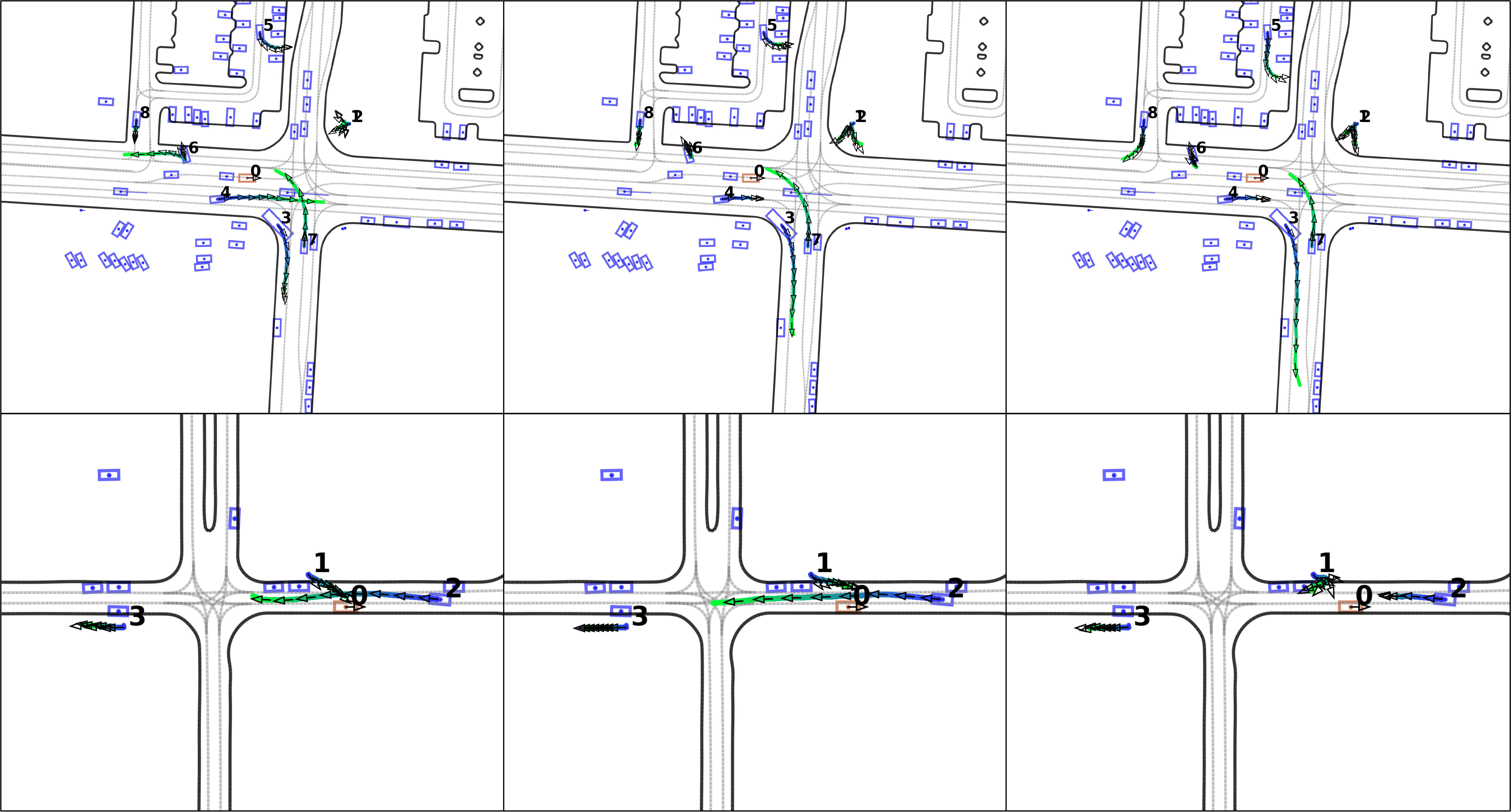}
    \caption{The results of three parallel rollouts after pre-training. \textbf{First row}: The model has learned the interaction among vehicles 3, 4, and 7 at the intersection. Interestingly, it has also learned multi-point U-turn behavior of vehicle 6, as well as parking spot selection behavior of vehicle 5. \textbf{Second row}: Vehicle 2 exhibits different behaviors depending on the predicted actions of pedestrian 1, including slow detouring, fast passing, and stopping to yield.}
    \label{fig:pretrain_qualitative_Results}
\end{figure}

Figure \ref{fig:pretrain_qualitative_Results} presents diverse rollout results generated by the model after pre-training. It can be observed that after extensive training, the model is able to simulate a variety of plausible future scenarios and generate reasonable future trajectories for different types of interactive behaviors.

\subsubsection{Post-training}
\begin{figure}[htbp]
    \centering
    \includegraphics[width=0.9\textwidth]{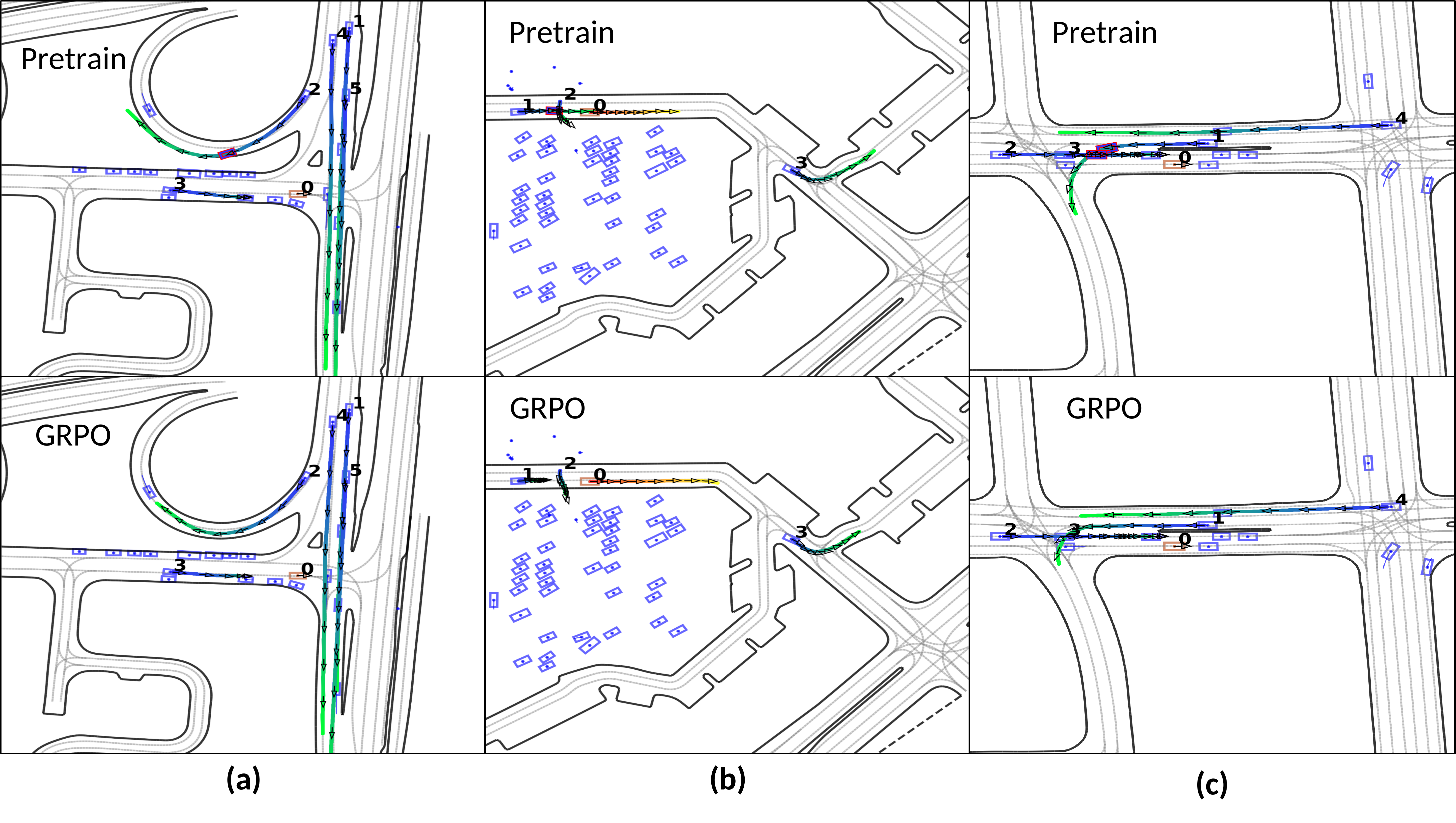}
    \caption{A qualitative comparison between GRPO and the pre-trained model. In (a), Vehicle 2 needs to perform a high-curvature turn to safely navigate the ramp. Such data is extremely rare and can be likened to the occurrence of consecutive uncommon words in large language models. Although the pre-trained policy shows some awareness of the need to turn, it still results in a collision. In contrast, the GRPO policy, despite reducing speed, is able to safely pass through the ramp. In (b), when encountering a pedestrian crossing the road, the pre-trained policy demonstrates some intention to yield but still ends up colliding, whereas the GRPO policy yields more thoroughly, ensuring safety. In (c), during an unprotected left turn scenario, the GRPO policy also exhibits greater ease and competence.}
    \label{fig:rl_qualitative_Results}
\end{figure}
Figure \ref{fig:rl_qualitative_Results} presents a comparison between the pre-trained policy and the GRPO policy in challenging scenarios under greedy sampling. It can be observed that although the pre-trained policy has learned a distribution similar to that of human behavior, it lacks sufficient causal reasoning in safety-critical situations, leading to potential safety risks. In contrast, the policy optimized through GRPO demonstrates the ability to generate safer motion plans when faced with safety-critical scenarios.

\clearpage
\subsection{Test-time Computing}
\begin{figure}[htbp]
    \centering
    \includegraphics[width=1.0\textwidth]{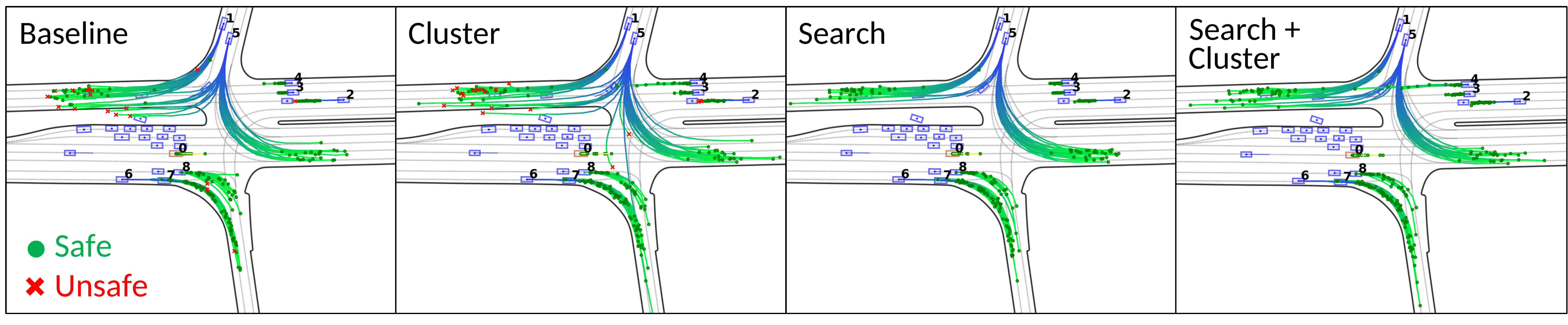}
    \caption{Qualitative comparison of test-time computing methods: Cluster vs. Search vs. Cluster+Search.}
    \label{fig:test_time_computing}
\end{figure}

We experimented with several test-time computing strategies, including cluster, search, and cluster+search. The results are shown in Figure \ref{fig:test_time_computing}. As illustrated, directly performing rollouts may produce unsafe trajectories due to inherent randomness. While clustering can enhance output diversity, it does not address the safety issue. The introduction of search fundamentally ensures the safety of the trajectories. Ultimately, the combination of search+cluster achieves a balance between safety and diversity, albeit at the cost of increased runtime.

\end{document}